\def\year{2021}\relax
\documentclass[letterpaper]{article} % DO NOT CHANGE THIS
\usepackage{aaai21}  % DO NOT CHANGE THIS
\usepackage{times}  % DO NOT CHANGE THIS
\usepackage{helvet} % DO NOT CHANGE THIS
\usepackage{courier}  % DO NOT CHANGE THIS
\usepackage[hyphens]{url}  % DO NOT CHANGE THIS
\usepackage{graphicx} % DO NOT CHANGE THIS
\usepackage{natbib}  % DO NOT CHANGE THIS AND DO NOT ADD ANY OPTIONS TO IT
\usepackage{caption} % DO NOT CHANGE THIS AND DO NOT ADD ANY OPTIONS TO IT
\usepackage{amsmath,amsfonts,bm}

\newcommand{\defuse}{{Defuse}\xspace}
\def\eqref#1{equation~\ref{#1}}
\def\1{\bm{1}}

\DeclareMathAlphabet{\mathsfit}{\encodingdefault}{\sfdefault}{m}{sl}
\SetMathAlphabet{\mathsfit}{bold}{\encodingdefault}{\sfdefault}{bx}{n}

\usepackage{xspace}

\usepackage{array}
\usepackage{hyperref}
\usepackage{url}
\usepackage{booktabs}       % professional-quality tables
\usepackage{amsmath,amsthm}
\usepackage{algorithmic, algorithm}
\usepackage{lipsum}
\usepackage{tabularx}
\usepackage{multirow}
\usepackage{graphicx,subfigure}

\newcommand{\hideh}[1]{}

\newif\ifcomments
\ifcomments
    \providecommand{\krishnaram}[2][]{{\protect\color{magenta}{[krishnaram:\textbf{#1} #2]}}}
    \providecommand{\nathalie}[2][]{{\protect\color{cyan}{[nathalie:\textbf{#1} #2]}}}
    \providecommand{\dylan}[2][]{{\protect\color{blue}{[dylan:\textbf{#1} #2]}}}
\else
    \providecommand{\krishnaram}[2][]{}
    \providecommand{\nathalie}[2][]{}
    \providecommand{\dylan}[2][]{}
\fi

\newtheorem{theorem}{Theorem}[section]
\newtheorem{definition}[theorem]{Definition}

\title{\defuse: Harnessing Unrestricted Adversarial Examples for Debugging Models Beyond Test Accuracy}
\author {
    Dylan Slack\textsuperscript{\rm 1,2},
    Nathalie Rauschmayr\textsuperscript{\rm 1}, 
    Krishnaram Kenthapadi\textsuperscript{\rm 1} \\
}
\affiliations {
    \textsuperscript{\rm 1} Amazon AWS AI, 
    \textsuperscript{\rm 2} University of California, Irvine\\
}

\begin{document}

\maketitle

\begin{abstract} 
We typically compute aggregate statistics on held-out test data to assess the generalization of machine learning models. However, statistics on test data often overstate model generalization, and thus, the performance of deployed machine learning models can be variable and untrustworthy. Motivated by these concerns, we develop methods to automatically discover and correct model errors beyond those available in the data. We propose \defuse, a method that generates novel model misclassifications, categorizes these errors into high-level ``model bugs'', and efficiently labels and fine-tunes on the errors to correct them.
To generate misclassified data, we propose an algorithm inspired by adversarial machine learning techniques that uses a generative model to find naturally occurring instances misclassified by a model. Further, we observe that the generative models have regions in their latent space with higher concentrations of misclassifications.  We call these regions \textit{misclassification regions} and find they have several useful properties. Each region contains a specific type of model bug; for instance, a misclassification region for an MNIST classifier contains a style of skinny $6$ that the model mistakes as a $1$.  We can also assign a single label to each region, facilitating low-cost labeling. We propose a method to learn the misclassification regions and use this insight to both categorize errors and correct them. In practice, \defuse finds and corrects novel errors in classifiers. For example, \defuse shows that a high-performance traffic sign classifier mistakes certain $50\textrm{km/h}$ signs as $80\textrm{km/h}$. \defuse corrects the error after fine-tuning while maintaining generalization on the test set.
% . We harness these methods to learn \textit{misclassification regions} in the latent space of generative models instead of individual adversarial examples. The misclassification  Further, the misclassification regions can be interpreted as specific model bugs, thus providing insight into how a model fails. We evaluate \defuse on high-performance image classifiers and find Defuse
% reveals critical and new model errors.
% Last, we show Defuse corrects the model errors while maintaining generalization according to
% the test set.
\end{abstract}

% \nathalie{Nathalie's macro for comments}
% \krishnaram{Krishnaram's macro for comments}

\section{Introduction}
\label{section:intro}

\begin{figure*}
    \centering
    \includegraphics[width=.85\textwidth]{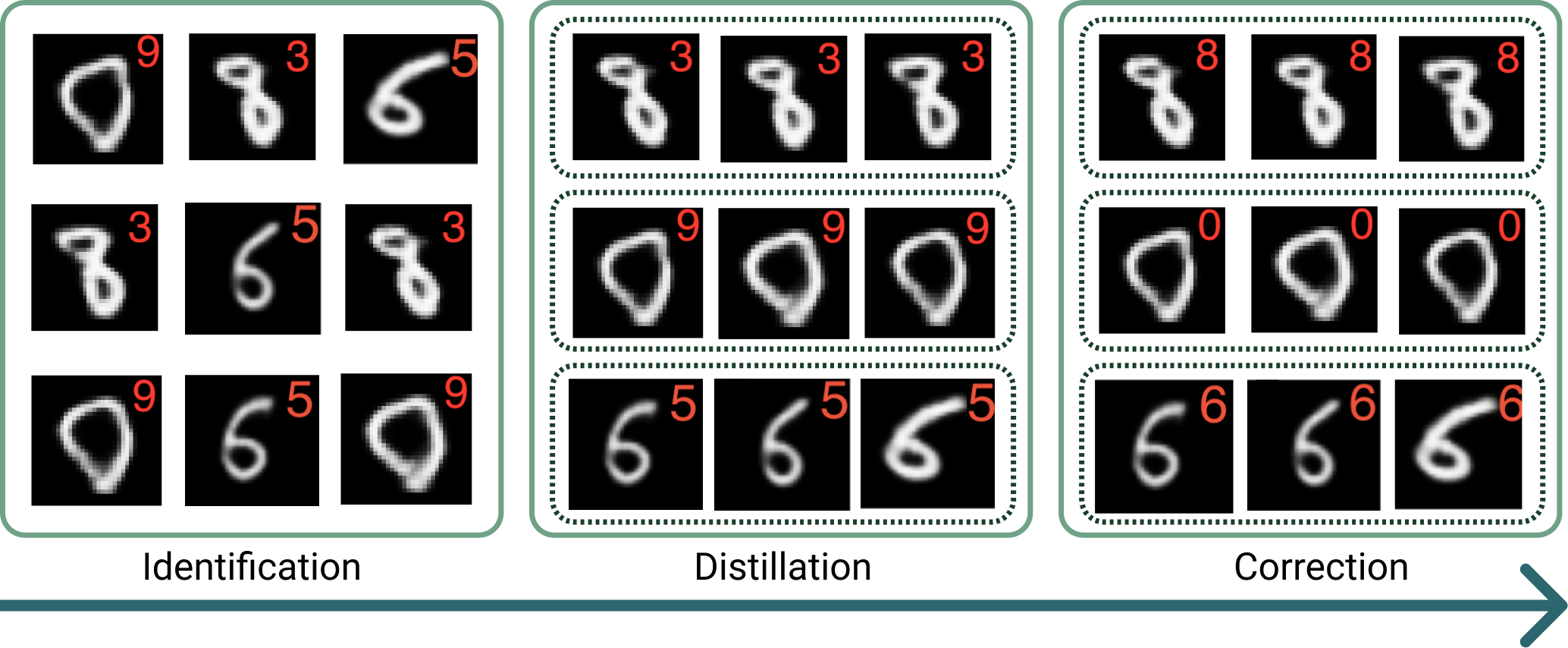}
    \vspace{-5mm}
    \caption{\textbf{Running \defuse on a MNIST classifier.} The (handpicked) images are examples from three misclassification regions identified from running \defuse.  The red digit in the upper right hand corner of the image is the classifier's prediction.  \defuse initially identifies many model failures.  Next, it aggregates these failures in the distillation step for annotator labeling.  Last, \defuse tunes the classifier so that it correctly classifies the images, with minimal change in classifier performance. \defuse serves as an end-to-end framework to diagnose and debug errors in classifiers.}
    \label{fig:MNIST-example}
\end{figure*}

%Discovering errors in machine learning (ML) models is a critical part of the ML development life cycle.  Uncovering bugs helps ML developers make important decisions about both development and deployment.  

%Machine learning models that generalize well according to held out test sets should ideally exhibit similar performance when deployed ``in the wild.'' Based on this principle, evaluation techniques often rely on aggregate test statistics such as held out accuracy. For instance, leader board style challenges are popular tool for tracking the best models on particular tasks [\cite{rajpurkar-etal-2016-squad}].  However, additional issues arise with over-reliance on test statistics. For instance, aggregate statistics like held out test accuracy are known to overestimate generalization performance [\cite{pmlr-v97-recht19a, ribeiro-etal-2020-beyond}]. 

A key goal of machine learning models is generalization.
We typically measure generalization through performance
on a held-out test set. Ideally, models that score well on
held-out test sets should perform the same when deployed.
Indeed, researchers track progress using leader boards that use such aggregate statistics [\cite{imagenet, rajpurkar-etal-2016-squad}]. Nevertheless, it has
become increasingly apparent test set accuracy alone does not
fully describe the performance of machine learning models. For
instance, statistics like held out test accuracy may overestimate generalization performance \cite{pmlr-v97-recht19a, ribeiro-etal-2020-beyond, kayurinvesting2008}. Also, test statistics offer little
insight into or remedy specific model failures \cite{wu-etal-2019-errudite}. Last, test data itself is often limited and may not cover all the possible deployment scenarios \cite{gardner-etal-2020-evaluating}. Because metrics on test set data often
fail to describe the performance of machine learning
systems fully, it is difficult to verify and trust the behavior of machine learning models when deployed.

%As a result, industry practitioners tend to adopt continuous evaluation and monitoring post deployment to understand whether the production data is significantly different than the training data and whether they can continue to trust  [\cite{sagemaker, modelmonitor}].   
As a result, researchers have developed a variety of techniques to evaluate models. Such methods include explanations \cite{ribeiro2016should,slack2020,shap-lunberg}, fairness metrics \cite{friedler-fairness, ji2020can}, and data set replication \cite{pmlr-v97-recht19a, Engstrom2020IdentifyingSB}.  Natural language processing (NLP) has increasingly turned to software engineering inspired behavioral testing tools to find errors in models \cite{ribeiro-etal-2020-beyond}.  Though these techniques may help find the reasons for misclassifications (e.g., explanations), they do not find novel situations in which the model fails.  Other routes to discover model errors are labor-intensive and may require a high amount of task-specific expertise (e.g., dataset replication and behavioral testing).  Separately, the adversarial machine learning literature has proposed techniques to generate misclassified inputs for machine learning models automatically.  Of particular interest, adversarial machine learning research proposes methods to generate naturally occurring inputs where humans and classifiers disagree called \textit{unrestricted} or \textit{natural} adversarial examples \cite{NIPS2018_8052, zhengli2018iclr}.  Instead of generating imperceptible perturbations like classic adversarial examples, unrestricted adversarial examples find semantically meaningful perturbations that cause models and humans to disagree. 
% These methods perturb an encoded representation of data instances in a generative model's latent space to produce on manifold perturbations. 
The adversarial example literature motivates these instances as a broader security threat than classic adversarial examples \cite{NIPS2018_8052}.  However, techniques that search for unrestricted adversarial examples find diverse instances where classifiers and humans disagree.  Thus, unrestricted adversarial examples as a general tool to determine model errors warrant further investigation.

%techniques that find misclassified instances are generally applicable to research areas 
%
%are inherently connected to research areas that develop methods aimed at finding model errors such as model evaluation and explainability. Thus, the connection warrants further investigation.

%What can this method do well --- explanations don't find specific model errors, fairness metrics assess something different than accuracy, data set replication + behavioral testing are extremely labor intensive

%
%However, these techniques do not provide methods to remedy model bugs or require a high level of human supervision.  To enable model designers to discover and correct model bugs beyond aggregate test statistics, we analyze \textit{unrestricted adversarial examples}: instances on the data manifold that are misclassified [\cite{NIPS2018_8052}]. We identify model bugs through diagnosing common patterns in unrestricted adversarial examples.

However, several issues arise when using unrestricted adversarial examples in model evaluation setting.  For one, unrestricted adversarial examples are generated as single instances and can be found in large quantity even for high-performance classifiers \cite{NIPS2018_8052, zhengli2018iclr}.  Uncovering actionable and general insights into model errors from a large set of mislabeled instances is highly challenging. Also, human annotators must verify the candidate samples as misclassifications imposing high costs. In this paper, we propose \textup{\defuse}: a method for \underline{de}bugging classi\underline{f}iers thro\underline{u}gh di\underline{s}tilling 
unr\underline{e}stricted adversarial examples.  \defuse facilities using unrestricted adversarial examples to discover and correct model errors while overcoming the issues mentioned earlier. Instead of focusing on single misclassified inputs, the framework provides general insights into model bugs through learning regions in the latent space with many similar errors.
  We call these regions \textit{misclassification regions}.  Also, \defuse uses the misclassification regions to label many unrestricted adversarial examples more efficiently.  All the instances in the region are similar and can receive the same label.  Thus, we only need to label the region and not every instance in the region.

For example, we run \defuse on a classifier trained on MNIST and provide an overview in figure \ref{fig:MNIST-example}. \defuse works in three steps: \textit{identification, distillation,} and \textit{correction}. In the identification step (first pane in figure \ref{fig:MNIST-example}), \defuse generates unrestricted adversarial examples.  The red number in the upper right-hand corner of the image is the classifier's prediction.  Although the classifier achieves high test set performance, we find naturally occurring examples that are classified incorrectly.  Next, the method performs the distillation step (second pane in figure \ref{fig:MNIST-example}). The clustering model groups together similar failures for annotator labeling.  For instance, \defuse groups together a certain type of incorrectly classified eight in the first row of the second pane in figure \ref{fig:MNIST-example}.  Next, \defuse receives annotator labels for each of the clusters.\footnote{We assign label $8$ to the first row in the second pane of figure \ref{fig:MNIST-example}, label $0$ to the second row, and label $6$ to the third row.} Last, we run the correction step using both the annotator labeled data and the original training data.  We see that the model correctly classifies the images (third pane in figure \ref{fig:MNIST-example}).  Importantly, the model maintains its predictive performance, scoring $99.1\%$ accuracy after tuning.  We see that \defuse serves as a general-purpose method to both discover and correct model errors. We highlight the following contributions of our work:

\begin{itemize}
\item We introduce \defuse, a method that can generate many novel model misclassifications through learning misclassification regions in the latent space of a generative model. 
\item We demonstrate that \defuse finds critical and novel bugs in a number of high performance image classifiers and verify the errors using human evaluation.  We also show that \defuse corrects the misclassifications without harming test set generalization. 
\item We demonstrate the general applicability of unrestricted adversarial examples as a tool for discovering model errors and their usefulness in learning misclassification regions. 
\end{itemize}

% \begin{figure}[h]
%     \centering
%     \includegraphics{}
%     \caption{Graphic illustrating training process}
%     \label{fig:example failure modes}
% \end{figure}

\section{Notation and Background}\label{sec:background}

In this section, we establish notation and background on unrestricted adversarial examples. Though unrestricted adversarial examples occur in many domains, we focus on \defuse applied to image classification.

\paragraph{Unrestricted adversarial examples} Let $f: \mathbb{R}^N \rightarrow [0, 1]^C$ denote a classifier that accepts a data point $x \in X$, where $X$ is the set of legitimate images.  The classifier $f$ returns the probability that $x$ belongs to class $c \in \{1,...,C\}$. Next, assume $f$ is trained on a data set $\mathcal{D}$ consisting of $d$ tuples $(x, y)$ containing data point $x$ and ground truth label $y$ using loss function $\mathcal{L}$. Finally, suppose there exists an oracle $o : x \in X \rightarrow \{1,...,C\}$ that outputs a label for $x$. 
We define unrestricted adversarial examples as the set $\mathcal{A}_N := \{ x \in X \textrm{ }| \textrm{ } o(x) \neq f(x)$\} [\cite{NIPS2018_8052}].  
%Then, we define an unrestricted adversarial example [\cite{NIPS2018_8052}] as the set $\mathcal{A}_N := \{ x \in X \textrm{ }| \textrm{ } o(x) \neq f(x)$\}.  This definition states unrestricted adversarial examples are the set of relevant images where the oracle and the classifier disagree.

\paragraph{Variational Autoencoders (VAEs)}  In order to find unrestricted adversarial examples, it is necessary to model the set of legitimate images $X$.  We use a VAE to create such a model. A VAE is composed of an encoder and a decoder neural networks.  These networks are used to model the relationship between data $x$ and latent factors $z \in \mathbb{R}^K$.  Where $x$ is generated by some ground truth  latent factors $v \in \mathbb{R}^M$, we wish to train a model such that the learned generative factors closely resemble the true factors: $p(x|v) \approx p(x|z)$. In order to train such a model, we employ the $\beta\textrm{-VAE}$ [\cite{Higgins2017betaVAELB}]. 
% $\beta\textrm{-VAE}$ encourages learning representations with independent latent factors $z$.
% Specifically, we impose isotropic unit Gaussian prior over the latent factors ($p(z)=\mathcal{N}(0,I)$) and establish parametric likelihood $p_\theta(x|z)$.  We train our model by minimizing the divergence between the data distribution and the model parameters with respect to $\theta$. Namely, we optimize $\argmin_\theta \textrm{KL} (p(x) || p_\theta(x)) = \argmax_\theta \mathbb{E}_{p(x)} [\textrm{log} p_\theta(x)]$ where $p_\theta (x) = \int p_\theta (x | z) p(z) dz$.  However, calculating $p_\theta (x)$ is intractable, so we introduce the parametric model $q_\phi(z|x)$ and construct the variational lower bound (ELBO) with respect to $\textrm{log } p_\theta(x)$, which can be shown to be:
% \begin{equation}
%     L(\theta, \phi; x, z) = \mathbb{E}_{q_\phi (z | x)}[\textrm{log} p_\theta(x|z)] - \beta \textrm{ } \textrm{KL}(q_\phi (z|x) || p(z))
%     \label{eq:b-vae-loss}
% \end{equation}
% We let the approximate posterior $q_\phi(z|x)$ take on Gaussian form with diagonal covariance.  Additionally, $\beta$ controls the strength of the constraint on the prior.  Increased $\beta$ generally encourages higher degrees of disentanglement in the latent representation [\cite{Higgins2017betaVAELB}].  
 This technique produces encoder $q_\phi(z|x)$ that maps from the data and latent codes and decoder $p_\theta(x|z)$ that maps from codes to data. 
%  The parameters of the neural networks defining the encoder and the decoder are $\phi$ and $\theta$ respectively. The parameters $\phi$ and $\theta$ are jointly optimized using stochastic gradient ascent.  

\hideh{
\paragraph{SSIM Loss} To improve the sample quality produced by $\beta$-VAE, we use the Structural Similarity Index (SSIM) [\cite{ssim_loss}, \cite{snell2017}]. We replace the reconstruction loss (i.e. first term) in equation \ref{eq:b-vae-loss} with the SSIM loss.  SSIM takes into account structure, luminance, and constrast to define similarity between two images. The SSIM is defined as: $\textrm{SSIM} = (2\mu_{x}\mu_{y} + C_{1})(\mu^{2}_{x}\mu^{2}_{y} + C_{1})^{-1} \cdot (2\sigma_{xy} + C_{2})(\sigma^{2}_{x} + \sigma^{2}_{y} + C_{2})^{-1}$. In this equation, $\mu_{x}$ and $\mu_{y}$ denote the mean pixel value and $\sigma_{y}$ and $\sigma_{x}$ denote the standard deviations. $C_{1}$, $C_{2}$, $C_{3}$ are constant values used for numerical stability. 
% SSIM takes into account the structure, luminance (expressed through the local means) and contrast (expressed through the standard deviations) of a set of pixels. 
This makes our objective: $
    L(\theta, \phi; x, z) = SSIM - \beta \textrm{ } \textrm{KL}(q_\phi (z|x) || p(z))$ }

\

\section{Methods}\label{sec:methods}

In this section, we introduce \defuse.  We describe the three main steps in the method: identification, distillation, and correction. We also formalize misclassification regions. 

\begin{figure}
    \centering
    \includegraphics[width=.42\textwidth]{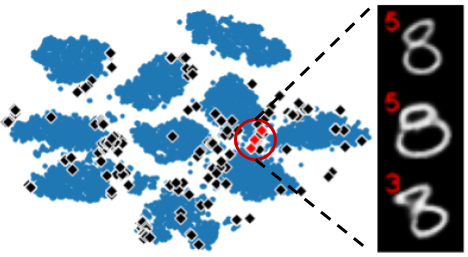}
    \caption{\textbf{Providing intuition for misclassification regions} through a t-SNE visualization of the latent space of MNIST.  The black diamonds correspond to the latent codes of unrestricted adversarial examples.  The blue circles are the latent codes of images from the training set. The images are three decoded latent codes (the red dots), where the red number in the left-hand corner is the classifier label.  We see that there are regions with higher densities of adversarial examples, }
    \label{fig:defusedistillationintuition}
\end{figure}

\subsection{Identification}
\label{subsec:identifyingfailurescenarios}

\begin{algorithm}[H]
\caption{Identification}
\label{alg:identifyingfailurescenarios}
\begin{algorithmic}[1]

\STATE {\bfseries Identify:} $f,p,q,x,y,a,b$
%\PROCEDURE{Identify}{$f,p,q,x,y,a,b$}       
    \STATE $\psi : = \{ \}$
    % \For{$x,y \in X,Y$}
    \STATE $\mu, \sigma : = q_\phi(x)$       
    \FOR{$i \in \{1,...,Q\}$}
        \STATE $\epsilon := [\textrm{Beta}(a,b)_1, ...,\textrm{Beta}(a,b)_M]$
        \STATE $x_{decoded} : = p_\theta(\mu + \epsilon)$
        \IF{$y \neq f(x_{decoded})$}   
            \STATE $\psi : = \psi \cup x_{decoded}$
        \ENDIF
    \ENDFOR
    % \EndFor
    \STATE Return $\psi$

\end{algorithmic}
\label{alg: indetification main}
\end{algorithm}

This section describes the \textit{identification} step in \defuse (first pane in figure \ref{fig:MNIST-example}). The aim of the \textit{identification} step is to generate many unrestricted adversarial examples for a model.
  We encode all the images from the training data. We perturb the latent codes with a small amount of noise drawn from a Beta distribution.  We use a Beta distribution so that it is possible to control the shape of the applied noise. We save instances that are classified differently from ground truth by the model $f$ when decoded. By perturbing the latent codes with a small amount of noise, we expect the decoded instances to have small but semantically meaningful differences from the original instances. Thus, if the classifier prediction deviates from the perturbation, the instance is likely misclassified. We denote the set of unrestricted adversarial examples for a single instance $\psi$.   We generate unrestricted adversarial examples over each instance $x \in X$, producing a set of unrestricted adversarial $\Psi$ containing the $\psi$ produced for each instance $x$. We provide pseudocode of the algorithm for generating unrestricted adversarial examples for a single instance $x$ in algorithm \ref{alg: indetification main}.
  
Our technique is related to the method from [\cite{zhengli2018iclr}]. The authors use a stochastic search method in the latent space of a GAN. They start with a small amount of noise and increase the noise's magnitude until they find an instance that is predicted differently than the original encoded instance.   Because we iterate over the entire data set, it is simpler to keep the noise fixed and sample a predetermined number of times. Critically, we save images that are predicted differently than the ground truth label of the original encoded instance and not just the original prediction. If the model misclassifies the original instance, we wish to save it as a model failure.  Otherwise, the method may not find errors associated with inputs that are misclassified incorrectly before perturbation. 

\subsection{Distillation}
\label{subsec:distillingfailurescenarios}

We first formalize misclassification regions. Next, we describe the distillation step: our procedure for learning the misclassification regions.  Recall, the misclassification regions are regions in the latent space of the generative model with high concentration of unrestricted adversarial examples. The regions provide insight into model bugs and can be used to efficiently correct errors.   

\subsubsection{Misclassification regions}

 Let $z \in \mathbb{R}^K$ be the latent codes corresponding to image $x \in X$ and $q_\phi(\cdot) : x \rightarrow z$ be the encoder mapping the relationship between images and latent codes.

\theoremstyle{definition}
\begin{definition}{\normalfont Misclassification region.}
Given a constant $\epsilon>0$, vector norm $||\cdot||$, model $f$, and point $z'$, a misclassification regions is a set of images $\mathcal{A}_R = \{ x \in X \textrm{ } | \textrm{ } \epsilon > ||q_\phi(x) - z'|| \land o(x) \neq f(x) \}$.
\end{definition}

Previous works that investigate unrestricted adversarial examples look for specific instances where the oracle and the model disagree [\cite{NIPS2018_8052, zhengli2018iclr}].  We instead look for regions in the latent space where this is the case.  Because the latent space of the VAE tends to take on Gaussian form due to the prior, we can use euclidean distance to define these regions.    If we were to define misclassification regions on the original data manifold, we may need a much more complex distance function.   Because it is likely too strict to assume the oracle and model disagree on \textit{every} instance in such a region, we also introduce a relaxation.

\theoremstyle{definition}
\begin{definition}{\normalfont Relaxed misclassification region.}
Given a constant $\epsilon>0$, vector norm $||\cdot||$, point $z'$, model $f$, and threshold $\rho$, a relaxed misclassification regions is a set of images $\mathcal{A}_f = \{ x \in X \textrm{ } | \textrm{ } \epsilon > ||q_\phi(x) - z'||\}$ such that $|\{ x \in \mathcal{A}_f \textrm{ } | \textrm{ } o(x) \neq f(x) \} | \textrm{ } / \textrm{ } |\mathcal{A}_f|  > \rho$.
\end{definition}

% This technique is very useful from a debugging perspective because it is possible to observe general model errors (i.e. a certain style of $8$ is mistaken as $3$ in figure \ref{fig:MNIST-example}) and provide a single label for the region. 
In this work, we adopt the latter definition of misclassification regions.  To concretize misclassification regions and provide evidence for their existence, we continue our MNIST example from figure \ref{fig:MNIST-example}. We plot the t-SNE embeddings of the latent codes of $10000$ images from the training set and $516$ unrestricted adversarial examples created during the identification step in figure \ref{fig:defusedistillationintuition} (details of how we generate unrestricted adversarial examples in section \ref{subsec:identifyingfailurescenarios}).  We see that the unrestricted adversarial examples are from similar regions in the latent space. 

\paragraph{Distilling misclassification regions}
Based on our definition of misclassification regions, we describe a general procedure for learning them.  We do so through clustering the latent codes of the unrestricted adversarial examples $\Psi$ in order to diagnose misclassification regions  (second pane of figure \ref{fig:MNIST-example}). 
We require our clustering method to (1) infer the correct number of clusters from the data, and (2) be capable of generating instances of each cluster.  We need to infer the number of clusters from the data because the number of misclassification regions is unknown ahead of time.  Further, we must generate many instances from each cluster so that we have enough data to finetune on to correct the faulty model behavior.  
Also, generating many failure instances enables model designers to see numerous examples from the misclassification regions, which encourages understanding the model failure modes.  Though any such clustering method under this description is compatible with distillation, we use a Gaussian mixture model (GMM) with the Dirichlet process prior.  We use the Dirichlet process because it describes the clustering problem where the number of mixtures is unknown beforehand, fulfilling our first criteria [\cite{ericsthesis}]. Additionally, because the model is generative, we can sample new instances, satisfying our second criteria.  

In practice, we use the truncated stick-breaking construction of the Dirichlet process, where $K$ is the upper bound of the number of mixtures. The truncated stick-breaking construction simplifies inference making computation more efficient [\cite{ericsthesis}].
The method outputs a set of clusters $\theta_j = (\mu_j, \sigma_j, \pi_j)$ where $j\in \{1,...,K\}$.  The parameters $\mu$ and $\sigma$ describe the mean and variance of a multivariate normal distribution and $\pi$ indicates the cluster weight.
To perform inference on the model, we employ expectation maximization (EM) described in [\cite{bishop2006}] and use the implementation provided in [\cite{scikit-learn}]. Once we run EM and determine the parameter values, we throw away cluster components that are not used by the model.  We fix some small $\epsilon$ and define the set of misclassification regions $\Lambda$ generated at the distillation step as: $\Lambda := \{ (\mu_j, \Sigma_j, \pi_j) | \pi_j > \epsilon \}$.

\begin{figure*}[ht]
\centering     %%% not \center
\subfigure{\label{fig:a}\includegraphics[width=40mm]{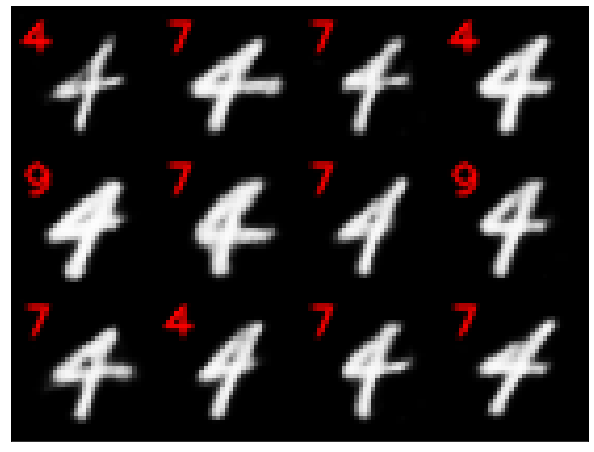}}
\subfigure{\label{fig:ab}\includegraphics[width=40mm]{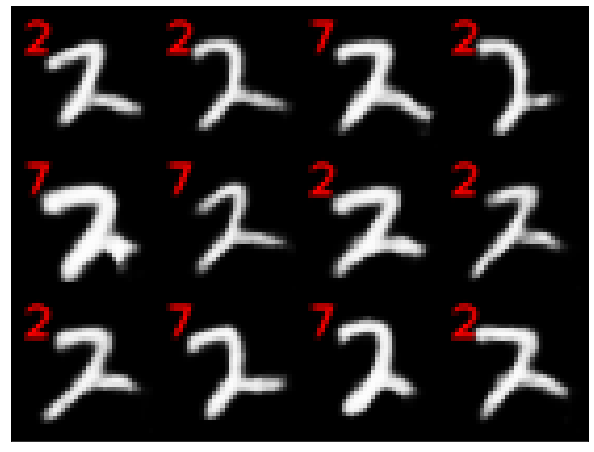}} 
\subfigure{\label{fig:ab}\includegraphics[width=40mm]{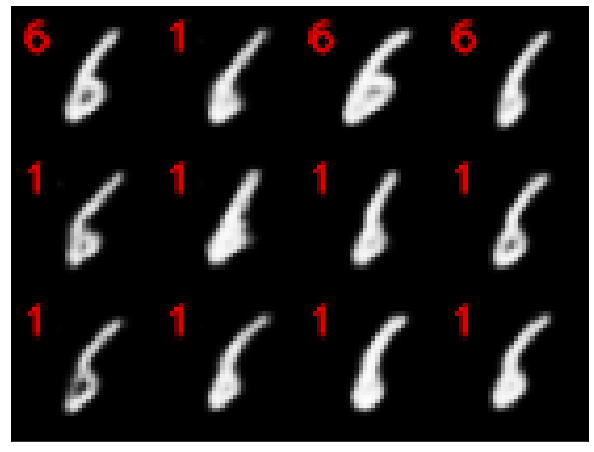}} 

\subfigure{\label{fig:svhmfailurescenario}\includegraphics[width=40mm]{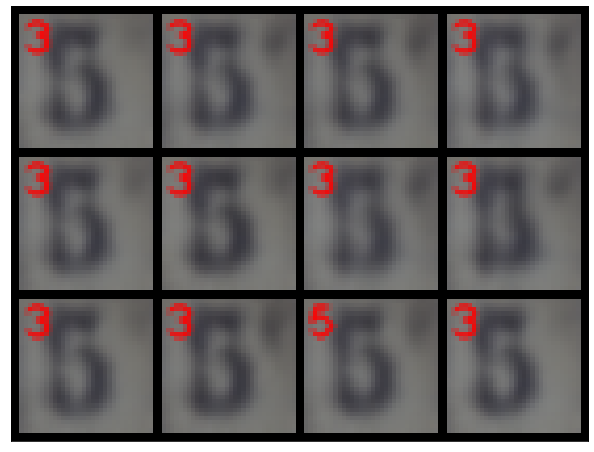}}
\subfigure{\label{fig:svhmfailurescenario2}\includegraphics[width=40mm]{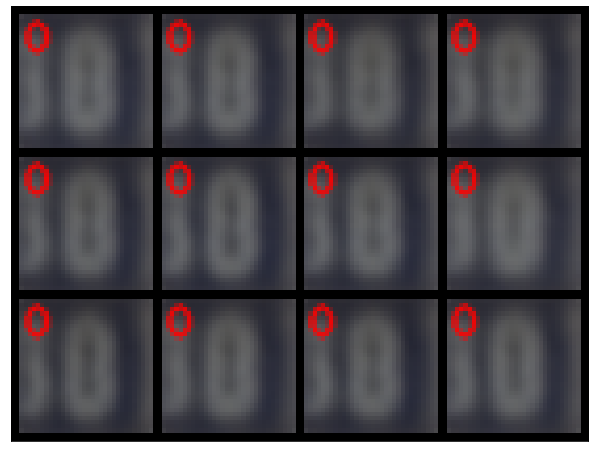}}
\subfigure{\label{fig:svhmfailurescenario2}\includegraphics[width=40mm]{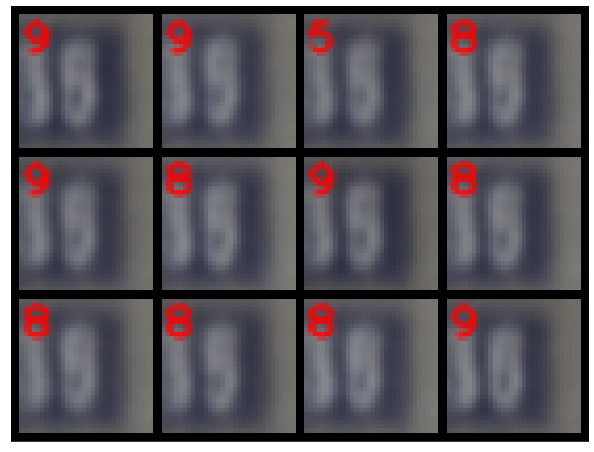}}

\subfigure{\label{fig:bb}\includegraphics[width=40mm]{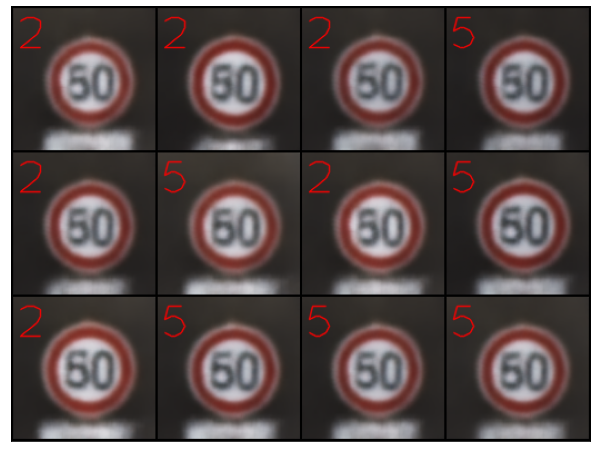}}
\subfigure{\label{fig:ba}\includegraphics[width=40mm]{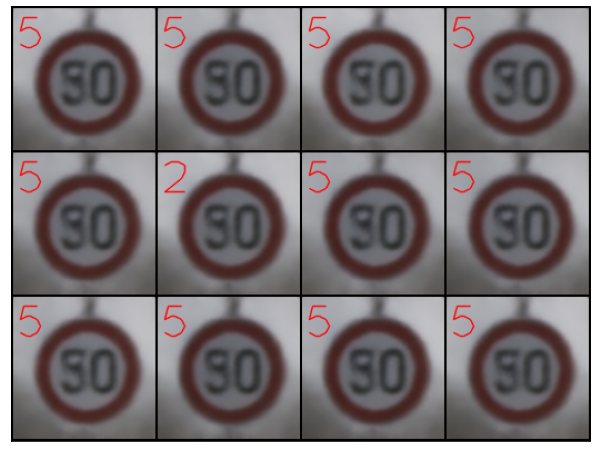}}
\subfigure{\label{fig:ba}\includegraphics[width=40mm]{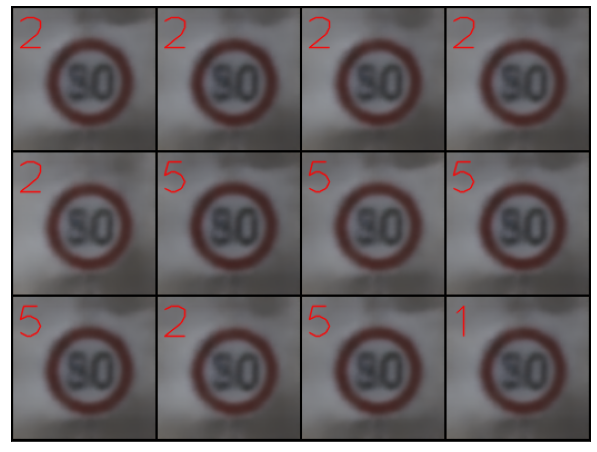}}
\caption{\textbf{Samples from three misclassification regions} from each dataset. \textbf{First row:} The MNIST misclassification regions. These scenarios were labeled $4$, $2$, $6$ in order from left to right. 
\textbf{Second row:} The SVHN misclassification regions labeled $5$, $8$, and $5$ from left to right.  \textbf{Third row:} The German signs misclassification regions. The label $1$ corresponds to $30$km/h, $2$ to $50$km/h, and $5$ to $80$km/h. The first and second were labeled $2$ while the third was labeled $1$.  \defuse finds significant bugs in the classifiers.
}
\label{fig:badsigns}
\end{figure*}

%% [KK] Duplicate text from next section
%%This section describes the procedure for labeling the misclassification regions and finetuning the model to fix the classifier errors.

\subsection{Correction}
\label{sec:correcting}
% This section describes the correction step in \defuse (third pane in figure \ref{fig:MNIST-example}).  In this step, an annotator assigns the correct label to the misclassification regions and tunes the model to resolve the mistakes.

This section describes the procedure for labeling the misclassification regions and finetuning the model to fix the classifier errors.

\paragraph{Labeling} First, an annotator assigns the correct label to the misclassification regions.  For each misclassification regions identified in $\Lambda$, we sample $Q$ latent codes from $z\sim \mathcal{N}(\mu_j, \tau \cdot \sigma_j)$.  Here, $\tau \in \mathbb{R}$ is a hyperparameter that controls sample diversity from the misclassification regions.  Because it could be possible for multiple ground truth classes to be present in a misclassification region, we set this parameter tight enough such that the sampled instances are from the same class. We reconstruct the latent codes using the decoder $p_\theta (x|z)$.  Next, an annotator reviews the reconstructed instances from the scenario and decides whether the scenario constitutes a model failure.  If so, the annotator assigns the correct label to all of the instances. The correct label constitutes a single label for all of the instances generated from the scenario.  We repeat this process for each of the scenarios identified in $\Lambda$ and produce a dataset of failure instances $\mathcal{D}_f$.  Pseudocode for the procedure is given in algorithm \ref{alg:label-failure-scenarios} in appendix \ref{sec:psuedo_code}. 

\paragraph{Finetuning} We finetune on the training data with an additional regularization term to fix the classifier performance on the misclassification regions.  The regularization term is the cross-entropy loss between the identified misclassification regions and the annotator label.  Where $\mathcal{L}$ is the cross-entropy loss applied to the failure instances $\mathcal{D}_f$ and $\lambda$ is the hyperparameter for the regularization term, we optimize the following objective using gradient descent,

\begin{align}
\mathcal{F}(\mathcal{D}, \mathcal{D}_f) =  \mathcal{L}(\mathcal{D}) + \lambda \cdot \mathcal{L}(\mathcal{D}_f)
\end{align}

This objective encourages the model to maintain its predictive performance on the original training data while encouraging the model to predict the failure instances correctly.  The regularization term $\lambda$ controls the pressure applied to the model to classify the failure instances correctly.

% \begin{figure}
% \centering     %%% not \center
% \subfigure{\label{fig:a}\includegraphics[width=45mm]{aies/images/mnistfailurescenarioexamples/example.png}}
% \subfigure{\label{fig:ab}\includegraphics[width=45mm]{aies/images/mnist2.png}} 
% \subfigure{\label{fig:ab}\includegraphics[width=45mm]{aies/images/mnist3.png}} 
% \subfigure{\label{fig:svhmfailurescenario}\includegraphics[width=45mm]{aies/images/svhn/svhn1.png}}
% \subfigure{\label{fig:svhmfailurescenario2}\includegraphics[width=45mm]{aies/images/svhn/svhn2.png}}
% \subfigure{\label{fig:svhmfailurescenario2}\includegraphics[width=45mm]{aies/images/svhn3.png}}
% \subfigure{\label{fig:bb}\includegraphics[width=45mm]{aies/images/signs_failure_scenarios/examplefailurescenario.png}}
% \subfigure{\label{fig:ba}\includegraphics[width=45mm]{aies/images/signs2.png}}
% \subfigure{\label{fig:ba}\includegraphics[width=45mm]{aies/images/signs3.png}}
% % \includegraphics[width=.75\textwidth]{aies/images/svhnexample.png}
% \caption{\textbf{Samples from three misclassification regions} from each dataset. \textbf{First row:} The MNIST misclassification regions. These scenarios were labeled $4$, $2$, $6$ in order from left to right. 
% \textbf{Second row:} The SVHN misclassification regions labeled $5$, $8$, and $5$ from left to right.  \textbf{Third row:} The German signs misclassification regions. The label $1$ corresponds to $30$km/h, $2$ to $50$km/h, and $5$ to $80$km/h. The first and second were labeled $2$ while the third was labeled $1$.  \defuse finds significant bugs in the classifiers.
% }
% \label{fig:badsigns}
% \end{figure}

\section{Experiments}\label{sec:expt}

% We first provide an illustrative example of \defuse using MNIST.  Then, we demonstrate \defuse on a traffic signs recognition task using the German traffic signs data set.

% In this section, we demonstrate the effectiveness of our approach on three data set: MNIST, the German traffic signs data set, and the street view house numbers data set.  
% We also run \defuse using NVAE (a large, state of the art VAE) on CIFAR-10 [\cite{vahdat2020NVAE}] to demonstrate its flexibility with other generative models. 

\hideh{
\begin{enumerate}
    \item We evaluate whether we identify model failures. 
    \item We sample a set of test images from the identified misclassification regions.  We assess the accuracy on this test set before and after correction.  After applying correction, the model should become more accurate on the scenarios.
    \item We assess test set performance before and after correction.  The model should maintain predictive performance after we correct on the misclassification regions. 
\end{enumerate}}
% Next, we describe our procedure for running \defuse on each data set and provide analysis.

\begin{figure*}
    \centering
    \begin{tabular}{llrrrrrrr} \toprule
        % & \multicolumn{2}{c}{Before finetuning} & \multicolumn{2}{c}{} \\
        & Dataset & \# Scenarios & Validation &  Test &  Misclassification Region  \\ \midrule
        \multirow{3}{*}{Before Finetuning} & MNIST & - & - & 98.3 & 29.1  \\
        & SVHN & - & 93.6 & 93.2 & 31.2 \\ \vspace{1mm}
        & German Signs & - & 98.8 & 98.7 & 27.8 \\
        Unrestricted & MNIST & - & - & 99.1 & 58.3   \\
        Adversarial Examples & SVHN & - & 93.1 & 92.9 & 65.4 \\ \vspace{1mm}
        & German Signs & - & - & - & - \\ 
        \multirow{3}{*}{\defuse} & MNIST & 19 & - & 99.1 & 96.4  \\
        & SVHN & 6 & 93.0 & 92.8 & 99.9 &  \\
        & German Signs & 8 & 98.1 &  97.7 & 85.6 \\
    \bottomrule
    \end{tabular}
    \caption{\textbf{Results from the best models} before finetuning, finetuning only on the unrestricted adversarial examples, and finetuning using \defuse.  The numbers presented are accuracy on the validation, test set, and misclassification region test set and the absolute number of misclassification regions generated using \defuse. We do not include finetuning on the unrestricted adversarial examples for German Signs because we, the authors, assigned misclassification regions for this data set and thus do not have ground truth labels for individual examples. Critically, the test accuracy on the misclassification regions is high for \defuse indicating that the method successfully corrects the faulty behavior.}
    \label{fig:tableresults}
\end{figure*}

\subsection{Setup}
\label{sec:setup}

\textbf{Datasets} We evaluate \defuse on three datasets: MNIST [\cite{lecun-mnisthandwrittendigit-2010}], the German Traffic Signs dataset [\cite{Stallkamp-IJCNN-2011}], and the Street view house numbers dataset (SVHN) [\cite{streetviewnumbers}].  MNIST consists of $60,000$  $32\textrm{X}32$ handwritten digits for training and $10,000$ digits for testing.  The images are labeled corresponding to the digits $0-9$.  
The German traffic signs data set includes $26,640$ training and $12,630$ testing images of size $128\textrm{X}128$.  We randomly split the testing data in half to produce a validation and testing set.  The images are labeled from $43$ different classes to indicate the type of traffic signs. The SVHN data set consists of $73,257$ training and $26,032$ testing images of size $32\textrm{X}32$. The images include digits of house numbers from Google streetview with labels $0-9$.  We split the testing set in half to produce a validation and testing set.

\textbf{Models} On MNIST, we train a CNN scoring $98.3\%$ test set accuracy following the architecture from [\cite{paszke2017automatic}]. 
On German traffic signs and SVHN, we finetune a Resnet18 model pretrained on ImageNet [\cite{He2016DeepRL}]. The German signs and SVHM models score $98.7\%$ and $93.2\%$ test accuracy respectively. 
We train a $\beta$-VAE the on the training data set to model the set of legitimate images in \defuse.  We use an Amazon EC2 P3 instance with a single NVIDIA Tesla V100 GPU for training.  We follow similar architectures to [\cite{Higgins2017betaVAELB}]. We set the size of the latent dimension $z$ to $10$ for MNIST/SVHN and $15$ for German signs. We provide our $\beta$-VAE architectures in appendix \ref{ap:trainingdetails}.

\textbf{Defuse}  We run \defuse on each classifier. In the identification step, we fix the  parameters of the Beta distribution noise $a$ and $b$ to $a=b=50.0$ for MNIST and $a=b=75.0$ for SVHN and German signs. We found these parameters were good choices because they produce a minimal amount of perturbation noise, making the decoded instances slightly different from the original instances.  
During distillation, we set the upper bound on the number of components $K$ to $100$.  We generally found the actual number of clusters to be much lower than this level.  Thus, this serves as an appropriate upper bound. We also fixed the weight threshold for clusters $\epsilon$ to $0.01$ during distillation to remove clusters with very low weighting. We also randomly downsample the number of unrestricted adversarial examples to $50,000$ to make the GMM more efficient.  
We sample finetuning and testing sets consisting of $256$ images each from every misclassification region for correction.  We found empirically that this number of samples is appropriate because it captures the breadth of possible images in the scenario.  We use the finetuning set as the set of failure instances $\mathcal{D}_f$.  We used the test set as held out data to evaluate classifier performance on the misclassification regions after correction.  During sampling, we fix the sample diversity $\tau$ to $0.5$ for MNIST and $0.01$ for SVHN and German signs because the samples from each of the misclassification regions appear to be in the same class using these values.  During correction, we finetune over a range of $\lambda$'s to find the best balance between training and misclassification region data.  We use $3$ epochs for MNIST and $5$ for both SVHN and German Signs because training converged within this amount of epochs.  During finetuning, we select the model for each $\lambda$ according to the highest training set accuracy for MNIST or validation set accuracy for SVHM and German traffic signs at the end of each finetuning epoch.  We select the best model overall as the highest training or validation performance over all $\lambda$'s.

\textbf{Annotator Labeling} Because \defuse requires human supervision, we use Amazon Sagemaker Ground Truth human workers to both determine whether clusters generated in the distillation step are misclassification regions and to generate their correct label.  To determine whether clusters are misclassification regions, we sample $10$ instances from each cluster in the distillation step. It is usually apparent the classifier disagrees with many of the ground truth labels within $10$ instances, and thus it is appropriate to label the cluster as a misclassification region. To reduce noise in the annotation process, we assign the same image to $5$ different workers and take the majority annotated label as ground truth.  The workers label the images using an interface that includes a single image and the possible labels for that task.  We additionally instruct workers to select ``None of the above'' if the image does not belong to any class and discard these labels.  For instance, the MNIST interface includes a single image and buttons for the digits $0-9$ along with a ``None of the above'' button. We provide a screenshot of this interface in figure \ref{fig:interface}. If more than half (i.e. setting $\rho=0.5$) of worker labeled instances disagree with the classifier predictions on the $10$ instances, we call the cluster a misclassification region.  We chose $\rho=0.5$ because clusters are highly dense with incorrect predictions at this level, making them useful for both understanding model failures and worthwhile for correction.  We take the majority prediction over each of the $10$ ground truth labels as the label for the misclassification region.   
As an exception, annotating the German traffic signs data requires specific knowledge of traffic signs.  The German traffic signs data ranges across $43$ different types of traffic signs.  It is not reasonable to assume annotators have enough familiarity with this data and can label it accurately.  For this data set, we, the authors, reviewed the distilled clusters and determined which clusters constituted misclassification regions. We labeled the clusters with more than half of the instances misclassified as misclassification regions. Though this procedure is less rigorous, the results still provide good insight into the model bugs discovered by \defuse.

\begin{figure*}[ht]
    \centering
    \subfigure[MNIST]{\label{fig:mnisttradeofff}\includegraphics[width=50mm]{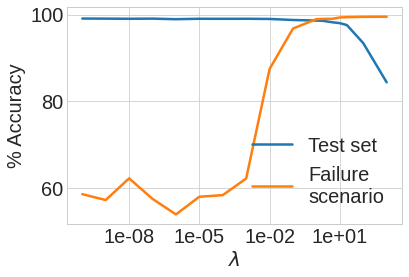}}
    \subfigure[SVHN]{\label{fig:svhntradeoff}\includegraphics[width=50mm]{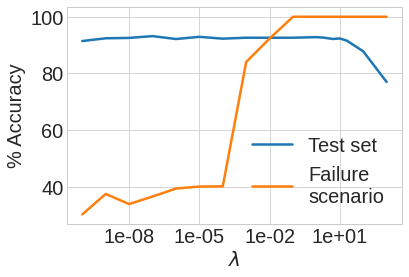}}
    \subfigure[German Signs]{\label{fig:germansignstradeoff} \includegraphics[width=50mm]{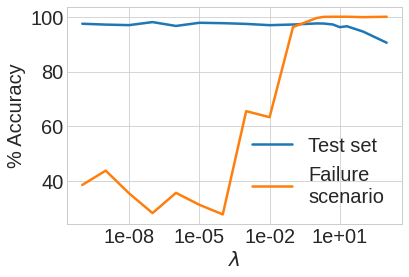}}
    % \begin{tabular}{lrrr} \toprule
    %     \bf Data set  &  Test set accuracy &  misclassification region accuracy \\ \midrule
    %     MNIST  & $98.3/99.1$ & $18.1/100.0$\\ 
    %     German Signs & $98.6/98.8$ & $89.8/100.0$ \\
    %     SVHN & $92.3/92.7$ & $0.0/42.5$ \\
    % \bottomrule
    % \end{tabular}
    \caption{\textbf{The tradeoff between test set and misclassification region accuracy} running correction. We assess both test set accuracy and accuracy on the test misclassification region data finetuning over a range of $\lambda$'s and plot the trade off.  There is an optimal $\lambda$ for each classifier where test set and misclassification region accuracy are both high. This result confirms that the correction step in \defuse adequately balances both generalization and accuracy on the misclassification regions . \dylan{adding the rest of the tradeoffs per bilal}}
    \label{fig:resultsoverview}
\end{figure*}

\subsection{Illustrative misclassification region examples}

We demonstrate \defuse's potential to identify critical model bugs.  We review misclassification regions from three datasets we consider.  \defuse returns $19$ misclassification regions for MNIST, $6$ for SVHN, and $8$ for German signs. We provide samples from three misclassification regions for each dataset in figure \ref{fig:badsigns}. The misclassification regions include numerous mislabeled examples of similar style. For example, in MNIST, the misclassification region in the upper left-hand corner of figure \ref{fig:badsigns} includes a similar style of incorrectly predicted $4$'s.  
% The same is true for the misclassification regions in the center and right column where a particular style of $2$'s and $6$'s are mistaken. 
The misclassification regions generally include ``corner case'' images.  These images are challenging to classify and thus highly insightful from a debugging perspective.  For instance, the misclassified $6$'s are relatively thin, making them appear like $1$'s in some cases. There are similar trends in SVHN and German Signs.  In SVHN, the model misclassifies particular types of $5$'s and $8$'s.  The same is true in German signs, where the model predicts styles of $50$km/h and $30$km/h signs incorrectly. We provide additional samples from other misclassification regions in appendix \ref{sec:additionalexperimentalresults}.  These results demonstrate \defuse uncovers significant and insightful model bugs. 

\subsection{Novely of the errors}

\begin{figure*}
    \centering
    \begin{tabular}{lr} \toprule
    Misclassified Training Set &  \includegraphics[width=100mm]{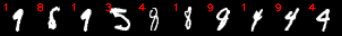} \\ 
    Misclassification Region & \includegraphics[width=100mm]{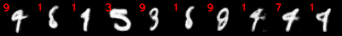}  \\\bottomrule
    \end{tabular}
    \caption{\textbf{Assessing the novelty of errors in  misclassification regions.} We compare samples from the misclassification regions with the nearest neighbors in the misclassified training set data.  We see that the misclassification regions reveal novel sources of model error not found in the misclassified training data.}
    \label{fig:errornovely}
\end{figure*}

We expect \defuse to find novel model misclassifications beyond those revealed by the available data. Thus, it is critical to evaluate whether the errors produced by \defuse are the same as those already in the training data. We compare the similarity of the errors proposed by \defuse (the misclassification region data) and the misclassified training data.  We perform this analysis on MNIST. We choose $10$ images from the misclassification regions and find the nearest neighbor in the misclassified training data according to $\ell_2$ distance.  We provide the results in figure \ref{fig:errornovely}.  We see that the data in the misclassification regions reveal different types of errors than the training set.

Interestingly, though some of the images are quite similar, they are predicted differently by the model.  This result indicates the misclassification regions reveal new model failures.  These results demonstrate \defuse can be used to reveal novel sources of model error. 

\subsection{Correcting misclassification regions}

After running correction, classifier accuracy should improve on the misclassification region data indicating we have corrected the bugs discovered in earlier steps. Also, the classifier accuracy on the test set should stay at a similar level or improve, indicating that the model generalization according to the test set is still strong.
We show that this is the case using \defuse. We assess accuracy on both the misclassification region test data and the original test set after performing correction.  We compare \defuse against finetuning only on the unrestricted adversarial examples labeled by annotators.  We expect this baseline to be reasonable because related works that focus on robustness to classic adversarial attacks demonstrate the effectiveness of tuning directly on the adversarial examples [\cite{pmlr-v97-zhang19p}].  We finetune on the unrestricted adversarial examples sweeping over a range of different $\lambda$'s and taking the best model as described in section \ref{sec:setup}. We use this baseline for MNIST and SVHN but not German Signs because we do not have ground truth labels for unrestricted adversarial examples.

We provide an overview of the models before finetuning, finetuning with the unrestricted adversarial examples, and using \defuse in figure \ref{fig:tableresults}. \defuse scores highly on the misclassification region data after correction compared to before finetuning.  There is only a marginal improvement in the baseline.  These results indicate \defuse corrects the faulty model performance on the misclassification regions. Also, these results show the clustering step in \defuse is critical to its success. We see this because finetuning on the unrestricted adversarial examples performs worse than finetuning on the misclassification regions.  Last, there are minor effects on test set performance during finetuning, demonstrating \defuse does not harm generalization according to the test set. 

Further, we plot the relationship between test set accuracy and misclassification region test accuracy in figure \ref{fig:resultsoverview} varying over $\lambda$.  There is an appropriate $\lambda$ for each model where test set accuracy and accuracy on the misclassification regions are both high.  Overall, these results indicate the correction step in \defuse is highly effective at correcting the errors discovered during identification and distillation.

\subsection{Annotator Agreement}

Because we rely on annotators to provide the ground truth labels for the unrestricted adversarial examples, we investigate the agreement between the annotators during labeling. The annotators should agree on the labels for the unrestricted adversarial examples.  Agreement indicates we have high confidence our evaluation is based on accurately labeled data. We evaluate the annotator agreement by assessing the percent of annotators that voted for the majority label prediction for a single instance.   This metric will be high when the annotators agree and low when only a few annotators constitute the majority vote. Further, we calculate the annotator agreement for every annotated instance. We provide the annotator agreement on MNIST and SVHN in figure \ref{fig:agreement} broken down into misclassification region data, non-misclassification region data, and their combination.

Interestingly, the misclassification region data has slightly lower annotator agreement indicating these tend to be more ambiguous examples. Further, there is lower agreement on SVHN than MNIST, likely because this data is more complex. All in all, there is generally high annotator agreement across all the data.

\begin{figure}
    \centering
    \begin{tabular}{lrrr} \toprule
    Dataset & M.R. & Non-M.R. & Combined \\  \midrule
    MNIST & $78.9.3\pm5.4$ & $87.2\pm3.2$ & $85.2\pm0.1$ \\
    SVHN & $66.6\pm8.4$ & $83.2\pm4.1$ & $82.1\pm1.3$ \\ \bottomrule
    \end{tabular}
    \caption{\textbf{Annotator agreement} on the unrestricted adversarial examples.  We plot the mean and standard error of the percent of annotators that voted for the majority label in an unresricted adversarial example across all the annotated examples.  We break this down into the misclassification region (M.R.) and non-misclassification region (Non-M.R.)  unrestricted adversarial examples and the combination between the two.  The annotators are generally in agreement though less so for the misclassification region data, indicating these tend to be more ambiguous examples.}
    \label{fig:agreement}
\end{figure}

\dylan{Experiments that could be good to add:
\begin{itemize}
    \item \# misclassification regions vs accuracy of the model
    \item Varying $\tau$
    \item Try and understand misclassification regions a bit more... what form do these regions take on
\end{itemize}}

\section{Related Work}

A number of related approaches for improving classifier performance use data created from generative models --- mostly generative adversarial networks (GANs) [\cite{Sandfort2019, Milz_2018_ECCV_Workshops,dataaugmentationgan}].  These methods use GANs to generate instances from classes that are underrepresented in the training data to improve generalization performance.  Additional methods use generative models for semi-supervised learning [\cite{NIPS2014_5352, Varma2016SocraticLA, NIPS2017_7137, advinference2016}].  Though these methods are similar in nature to the correction step of our work, a key difference is \defuse focuses on summarizing and presenting high level model failures.
% \defuse highlights high level model failures --- mistaking cats and deer for instances.  
% These are useful from both an interpretability perspective (i.e. understanding the behavior of the model) as well as fixing errors in the model. 
% Additional methods use GAN's to generate natural adversarial examples. For instance, both [\cite{zhengli2018iclr}] and  [\cite{NIPS2018_8052}] search over the latent space of GANs to find naturally occurring adversarial examples.  Our methods build on these works but differ by using VAE's, a novel search strategy, and focus on using clustered natural adversarial examples to correct classifier bugs.
% Related, [\cite{explainingmistakes}] use the latent space of a GAN to determine semantic changes to an image such that the class changes
Also, [\cite{flipper}] provide a system to debug data generated from a GAN when the training set may be inaccurate.  Though similar, we ultimately use a generative model to debug a classifier and do not focus on the generative model itself.  Last, similar to [\cite{NIPS2018_8052}, \cite{zhengli2018iclr}], [\cite{boothbayestrex}] provide a method to generate highly confident misclassified instances.

One aspect of our work looks at improving performance on unrestricted adversarial examples. Thus, there are similarities between our work and methods that improve robustness to adversarial attacks. Similar to \defuse, several techniques demonstrate that tuning on additional data helps improve classic adversarial robustness.  \cite{carmon2019unlabeled} demonstrate robustness is improved with the addition of unlabeled data during training. Also, \cite{Zhang2018TrainingSD} show directly training on the adversarial examples improves robustness. \cite{pmlr-v119-raghunathan20a} characterize a trade-off between robustness and accuracy in perturbation based data augmentations during adversarial training.  Because we train with on manifold data created by generative models, there is not so much of a trade-off between robustness and accuracy. We find that we can achieve high performance on the unrestricted adversarial examples with minimal change to test accuracy. Though related, our work demonstrates that robustness to naturally occurring adversarial examples show different robustness dynamics than classic adversarial examples.   

Related to debugging models, [\cite{Kang2018ModelAF}] focus on model assertions that flag failures during production. 
Also, [\cite{Zhang2018TrainingSD}] investigate debugging the training set for incorrectly labeled instances. We focus on preemptively identifying model bugs and do not focus on incorrectly labeled test set instances. Additionally, [\cite{ribeiro-etal-2020-beyond}] propose a set of behavioral testing tools that help model designers find bugs in NLP models. This technique requires a high level of supervision and thus might not be appropraite in some settings. Last, [\cite{pmlr-v97-odena19a}] provide a technique to debug neural networks through perturbing data inputs with various types of noise.  By leveraging unrestricted adversarial examples, we distill high level patterns in critical and naturally occurring model bugs. This technique requires minimal human supervision while presenting important types of model errors to designers. 
% Our method identifies with only minimal task knowledge. 

\section{Conclusion}

% \dylan{strong ending, maybe add bullet points if space needed in introduction with 3-4 bullet points of contributions, add with better ending ideas, bit more summary, split into two paragraphs, where one paragraph is conclusion and other is the future directions / discussin}

In this paper, we present \defuse: a method that generates and aggregates unrestricted adversarial examples to debug classifiers.  Though previous works discuss unrestricted adversarial examples, we harness such examples to debug classifiers.  We accomplish this task by identifying misclassification regions: regions in the latent space of a VAE with many unrestricted adversarial examples. On various data sets, we find that samples from misclassification regions are useful in many ways.  First, misclassification regions are informative for understanding the ways specific models fail.  Second, the generative aspect of misclassification regions is beneficial for correcting misclassification regions. Our experimental results show that these misclassification regions include critical model issues for classifiers with real world impacts (i.e. traffic sign classification) and verify our results using ground truth annotator labels. We demonstrate that \defuse successfully resolves these issues.  A potential direction for future research is to explore directly optimizing for the misclassification regions instead of using unrestricted adversarial examples.  Another direction worth exploring is to evaluate the success of \defuse applied to large, state of the art generative models.

\bibliography{bibliography}

\clearpage
\appendix

\section{\defuse Psuedo Code}
\label{sec:psuedo_code}

In algorithm \ref{alg:label-failure-scenarios}, $\texttt{Correct}(\cdot)$ and $\texttt{Label}(\cdot)$ are the steps where the annotator decides if the scenario warrants correction and the annotator label for the misclassification region. 

\begin{minipage}{0.46\textwidth}
\begin{algorithm}[H]
\caption{Identification}
\label{alg:identifyingfailurescenarios}
\begin{algorithmic}[1]

\STATE {\bfseries Identify:} $f,p,q,x,y,a,b$
%\PROCEDURE{Identify}{$f,p,q,x,y,a,b$}       
    \STATE $\psi : = \{ \}$
    % \For{$x,y \in X,Y$}
    \STATE $\mu, \sigma : = q_\phi(x)$       
    \FOR{$i \in \{1,...,Q\}$}
        \STATE $\epsilon := [\textrm{Beta}(a,b)_1,$
        \STATE \hspace{8mm} $...,\textrm{Beta}(a,b)_M]$
        \STATE $x_{decoded} : = p_\theta(\mu + \epsilon)$
        \IF{$y \neq f(x_{decoded})$}   
            \STATE $\psi : = \psi \cup x_{decoded}$
        \ENDIF
    \ENDFOR
    % \EndFor
    \STATE Return $\psi$

\end{algorithmic}
\end{algorithm}
\end{minipage}
\hfill
\begin{minipage}{0.46\textwidth}
\begin{algorithm}[H]
\caption{Labeling}
\label{alg:label-failure-scenarios}
\begin{algorithmic}
\STATE{Label Scenarios} $Q,\Lambda,p,q, \tau$   
    \STATE $D_f : = \{ \}$
    \FOR{$(\mu,\sigma, \pi) \in \Lambda$}
        \STATE $X_{d} : = \{ \}$
        \FOR{$i \in \{1,..,Q\}$}
            \STATE $X_{d} : = X_{d} \cup q_\psi(\mathcal{N}(\mu, \tau \cdot \sigma))$
        \ENDFOR
        \IF{\texttt{Correct}$(X_{d})$}   
            \STATE $D_f : = D_f  \cup \{ X_{d}, \textrm{\texttt{Label}}(X_{d}) \}$
        \ENDIF
    \ENDFOR
    % \EndFor
    \STATE Return $\bigcup D_f$
% \EndProcedure
\end{algorithmic}
\end{algorithm}
\end{minipage}

\section{Training details}
\label{ap:trainingdetails}

\subsection{GMM details}
\label{ap:gmmdetails}

In all experiments, we use the implementation of Gaussian mixture model with dirichlet process prior from [\cite{scikit-learn}].  We run our experiments with the default parameters and full component covariance.  

\subsection{MNIST details}
\label{ap:mnist-details}

\paragraph{Model details} We train a CNN on the MNIST data set using the architecture in figure \ref{tab:mnistarchitecutre}.  We used the Adadelta optimizer with the learning rate set to $1$.  We trained for $5$ epochs with a batch size of $64$.  

\begin{figure}[h!]
    \begin{tabular}{|l|}
    \hline
    Architecture                  \\ \hline
    4x4 conv., 64 ReLU stride 2   \\ \hline
    4x4 conv., 64 ReLU stride 2   \\ \hline
    4x4 conv., 64 ReLU stride 2   \\ \hline
    4x4 conv., 64 ReLU stride 2   \\ \hline
    Fully connected 256, ReLU           \\ \hline
    Fully connected 256, ReLU           \\ \hline
    Fully connected 10 $\times$ 2 \\ \hline
    \end{tabular}
    \centering
    \caption{MNIST CNN Architecture}
    \label{tab:mnistarchitecutre}
\end{figure}

\paragraph{$\beta$-VAE training details} We train a $\beta$-VAE on MNIST using the architectures in figure \ref{tab:encMNIST} and \ref{tab:decMNIST}.  We set $\beta$ to $4$. We trained for $800$ epochs using the Adam optimizer with a learning rate of $0.001$, a minibatch size of $2048$, and $\beta$ set to $0.4$.  We also applied a linear annealing schedule on the KL-Divergence for $500$ optimization steps. We set $z$ to have $10$ dimensions. 

\begin{figure}[h!]
    \begin{tabular}{|l|}
    \hline
    Architecture                  \\ \hline
    4x4 conv., 32 ReLU stride 2   \\ \hline
    4x4 conv., 32 ReLU stride 2   \\ \hline
    4x4 conv., 32 ReLU stride 2   \\ \hline
    Fully connected 256, ReLU           \\ \hline
    Fully connected 256, ReLU           \\ \hline
    Fully connected 15 $\times$ 2 \\ \hline
    \end{tabular}
    \centering
    \caption{MNIST data set encoder architecture.}
    \label{tab:encMNIST}
\end{figure}

\begin{figure}[h!]
    \begin{tabular}{|l|}
    \hline
    Architecture                  \\ \hline
    Fully connected 256, ReLU           \\ \hline
    Fully connected 256, ReLU           \\ \hline
    Fully connected 256, ReLU           \\ \hline
    4x4 transpose conv., 32 ReLU stride 2   \\ \hline
    4x4 transpose conv., 32 ReLU stride 2   \\ \hline
    4x4 transpose conv., 32 ReLU stride 2   \\ \hline
    4x4 transpose conv., 32 Sigmoid stride 2   \\ \hline
    \end{tabular}
    \centering
    \caption{MNIST data set decoder architecture.}
    \label{tab:decMNIST}
\end{figure}

\paragraph{Identification} We performed identification with $Q$ set to $500$.  We set $a$ and $b$ both to $50$.   We ran identification over the entire training set. Last, we limited the max allowable size of $\psi$ to $100$.  

\paragraph{Distillation} We ran the distillation step setting $K$, the upper bound on the number of mixtures, to $100$.  We fixed $\epsilon$ to $0.01$ and discarded clusters with mixing proportions less than this value.  This left $44$ possible scenarios.  We set $\tau$ to $0.5$ during review. We used Amazon Sagemaker Ground Truth to determine misclassification regions and labels.  The labeling procedure is described in section \ref{sec:setup}. This produced $19$ misclassification regions.  

\paragraph{Correction} We sampled $256$ images from each of the misclassification regions for both finetuning and testing.  We finetuned with minibatch size of $256$, the Adam optimizer, and learning rate set to $0.001$.  We swept over a range of correction regularization $\lambda$'s consisting of $[1e-10,1e-9,1e-8,1e-7,1e-6,1e-5,1e-4,1e-3,1e-2,1e-1,1,2,5,10,20,100,1000]$ and finetuned for $3$ epochs on each.

\subsection{German Signs Dataset Details}
\label{ap:architecture}

\paragraph{Dataset} The data consists of $26640$ training images and $12630$ testing images consisting of 43 different types of traffic signs. 
We randomly split the testing data in half to produce $6315$ testing and validation images. 
Additionally, we resize the images to $128\textrm{x}128$ pixels.

\paragraph{Classifier $f$} We fine-tuned the ResNet18 model for $20$ epochs using Adam with the cross entropy loss, learning rate of $0.001$, batch size of $256$ on the training data set, and assessed the validation accuracy at the end of each epoch.  We saved the model with the highest validation accuracy.  

\paragraph{$\beta$-VAE training details} We trained for $800$ epochs using the Adam optimizer with a learning rate of $0.001$, a minibatch size of $2048$, and $\beta$ set to $4$.  We also applied a linear annealing schedule on the KL-Divergence for $500$ optimization steps.  We set $z$ to have $15$ dimensions. 

\begin{figure}[h!]
    \begin{tabular}{|l|}
    \hline
    Architecture                  \\ \hline
    4x4 conv., 64 ReLU stride 2   \\ \hline
    4x4 conv., 64 ReLU stride 2   \\ \hline
    4x4 conv., 64 ReLU stride 2   \\ \hline
    4x4 conv., 64 ReLU stride 2   \\ \hline
    Fully connected 256, ReLU           \\ \hline
    Fully connected 256, ReLU           \\ \hline
    Fully connected 15 $\times$ 2 \\ \hline
    \end{tabular}
    \centering
    \caption{German signs data set encoder architecture.}
    \label{tab:encodersigns}
\end{figure}

\begin{figure}[h!]
    \begin{tabular}{|l|}
    \hline
    Architecture                  \\ \hline
    Fully connected 256, ReLU           \\ \hline
    Fully connected 256, ReLU           \\ \hline
    Fully connected 256, ReLU           \\ \hline
    4x4 transpose conv., 64 ReLU stride 2   \\ \hline
    4x4 transpose conv., 64 ReLU stride 2   \\ \hline
    4x4 transpose conv., 64 ReLU stride 2   \\ \hline
    4x4 transpose conv., 64 ReLU stride 2   \\ \hline
    4x4 transpose conv., 64 Sigmoid stride 2   \\ \hline
    \end{tabular}
    \centering
    \caption{German signs data set decoder architecture.}
    \label{tab:decsigns}
\end{figure}

\paragraph{Identification} We performed identification with $Q$ set to $100$.  We set $a$ and $b$ both to $75$.

\paragraph{Distillation} We ran the distillation step setting $K$ to $100$.  We fixed $\epsilon$ to $0.01$ and discarded clusters with mixing proportions less than this value.  This left $38$ possible scenarios.  We set $\tau$ to $0.01$ during review. We determined $8$ of these scenarios were particularly concerning.

\paragraph{Correction} We finetuned with minibatch size of $256$, the Adam optimizer, and learning rate set to $0.001$.  We swept over a range of correction regularization $\lambda$'s consisting of $[1e-10,1e-9,1e-8,1e-7,1e-6,1e-5,1e-4,1e-3,1e-2,1e-1,1,2,5,10,20,100,1000]$ and finetuned for $5$ epochs on each. 

\subsection{SVHN details}
\label{ap:svhnarchitectures}

\paragraph{Dataset} The data set consists of  $73257$ training and $26032$ testing images. We also randomly split the testing data to create a validation data set.  Thus, the final validation and testing set correspond to $13016$ images each.

\paragraph{Classifier $f$} We fine tuned for $10$ epochs using the Adam optimizer, learning rate set to $0.001$, and a batch size of $2048$.  We chose the model which scored the best validation accuracy when measured at the end of each epoch.  

\paragraph{$\beta$-VAE training details} We trained the $\beta$-VAE for $400$ epochs using the Adam optimizer, learning rate $0.001$, and minibatch size of $2048$. We set $\beta$ to $4$ and applied a linear annealing schedule on the Kl-Divergence for $5000$ optimization steps.  We set $z$ to have $10$ dimensions. 

\begin{figure}[h!]
    \begin{tabular}{|l|}
    \hline
    Architecture                  \\ \hline
    4x4 conv., 64 ReLU stride 2   \\ \hline
    4x4 conv., 64 ReLU stride 2   \\ \hline
    4x4 conv., 64 ReLU stride 2   \\ \hline
    Fully connected 256, ReLU           \\ \hline
    Fully connected 256, ReLU           \\ \hline
    Fully connected 10 $\times$ 2 \\ \hline
    \end{tabular}
    \centering
    \caption{SVHN data set encoder architecture.}
    \label{tab:encSVHN}
\end{figure}

\begin{figure}[h!]
    \begin{tabular}{|l|}
    \hline
    Architecture                  \\ \hline
    Fully connected 256, ReLU           \\ \hline
    Fully connected 256, ReLU           \\ \hline
    Fully connected 256, ReLU           \\ \hline
    4x4 transpose conv., 64 ReLU stride 2   \\ \hline
    4x4 transpose conv., 64 ReLU stride 2   \\ \hline
    4x4 transpose conv., 64 ReLU stride 2   \\ \hline
    4x4 transpose conv., 64 Sigmoid stride 2   \\ \hline
    \end{tabular}
    \centering
    \caption{SVHN data set decoder architecture.}
    \label{tab:decSVHN}
\end{figure}

\paragraph{Identification} We set $Q$ to $100$.  We also set the maximum size of $\psi$ to $10$. We set $a$ and $b$ to $75$.

\paragraph{Distillation}  We set $K$ to $100$.  We fixed $\epsilon$ to $0.01$.  The distillation step identified $32$ plausible misclassification regions.  The annotators deemed $6$ of these to be misclassification regions.  We set $\tau$ to $0.01$ during review. 

\paragraph{Correction} We set the minibatch size of $2048$, the Adam optimizer, and learning rate set to $0.001$.  We considered a range of $\lambda$'s: $[1e-10,1e-9,1e-8,1e-7,1e-6,1e-5,1e-4,1e-3,1e-2,1e-1,1,2,5,10,20,100,1000]$.  We finetuned for $5$ epochs.

\subsection{t-SNE example details}

We run t-SNE on $10,000$ examples from the training data and $516$ unrestricted adversarial examples setting perplexity to $30$. For the sake of clarity, we do not include outliers from the unrestricted adversarial examples.  Namely, we only include unrestricted adversarial examples with $>1\%$ probabilitdefuse-iclr clusters.

\section{Annotator interface}

\begin{figure*}
    \centering
    \includegraphics[width=\textwidth]{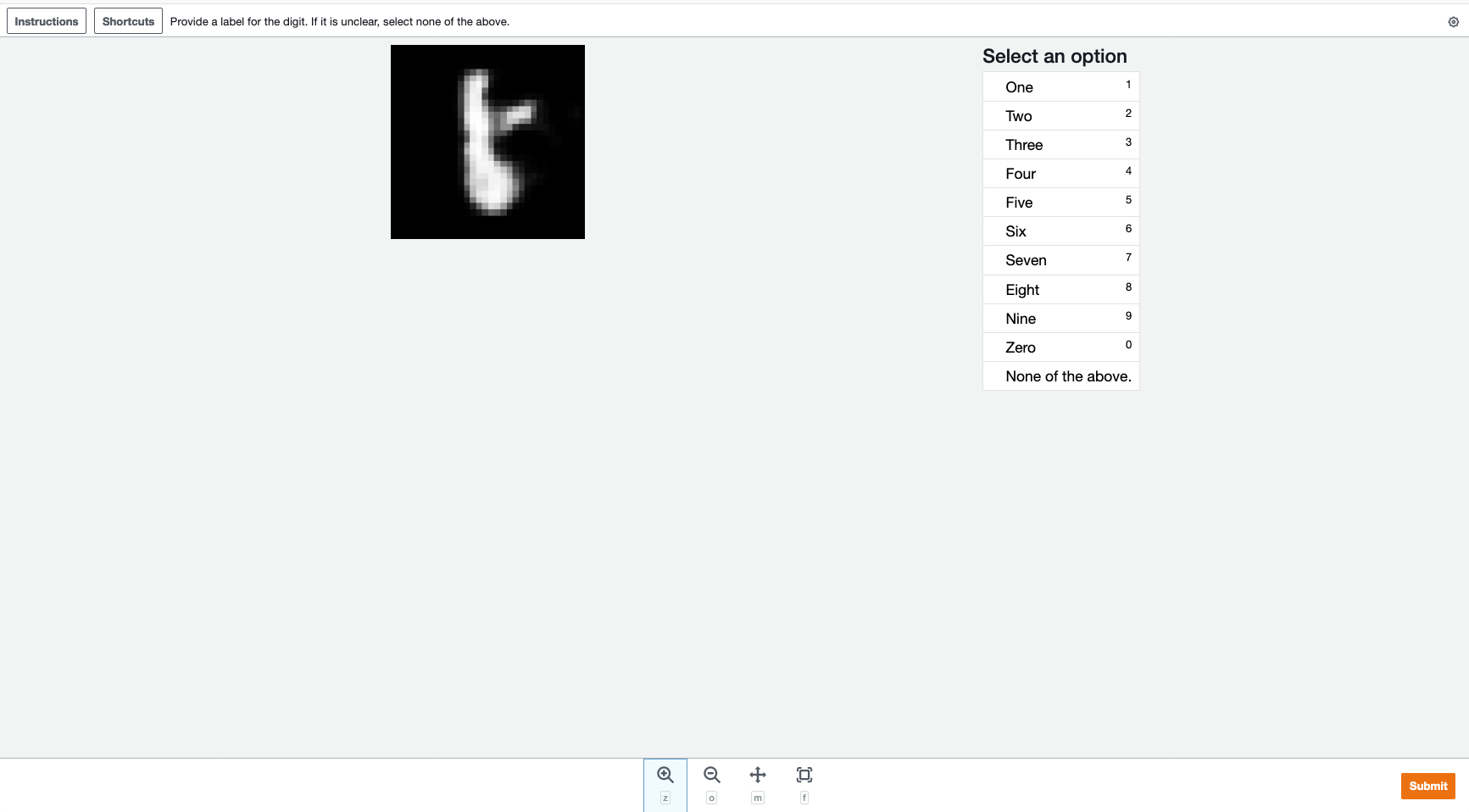}
    \caption{Annotation interface.}
    \label{fig:interface}
\end{figure*}

We provide a screenshot of the annotator interface in figure \ref{fig:interface}.

\section{Additional experimental results}
\label{sec:additionalexperimentalresults}

\subsection{Additional samples from MNIST misclassification regions}
\label{ap:additionmnist}

We provide additional examples from $10$ randomly selected (no cherry picking) MNIST misclassification regions. We include the annotator consensus label for each misclassification region.

\begin{figure*}[h!]
    \centering
    \includegraphics[width=.5\textwidth]{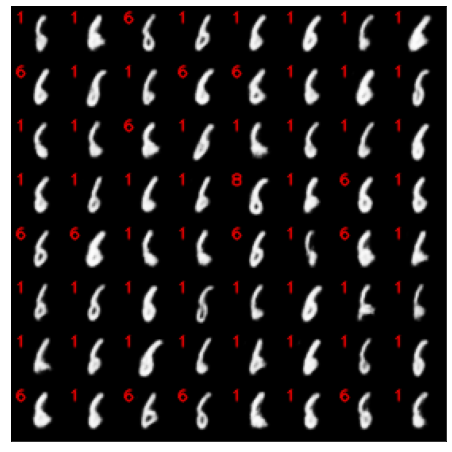}
    \caption{Annotator label $6$.}
\end{figure*}

% \begin{figure*}[h!]
%     \centering
%     \includegraphics[width=.5\textwidth]{images/mnistfailurescenarioexamples/21-8.png}
%     \caption{Annotator label $8$.}
% \end{figure*}

\begin{figure*}[h!]
    \centering
    \includegraphics[width=.5\textwidth]{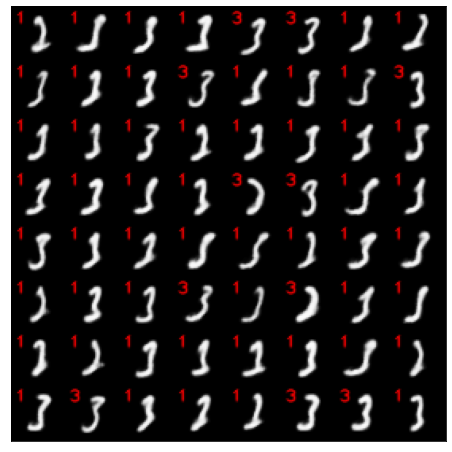}
    \caption{Annotator label $3$.}
\end{figure*}
\begin{figure*}[h!]
    \centering
    \includegraphics[width=.5\textwidth]{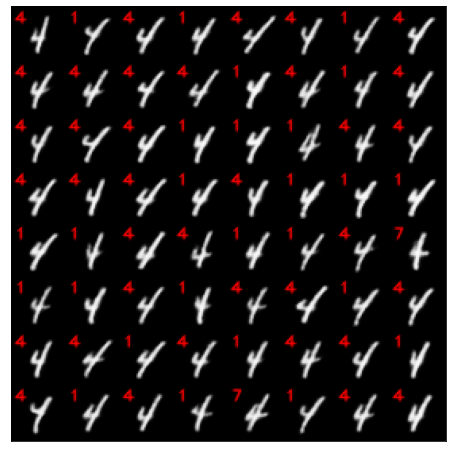}
    \caption{Annotator label $4$.}
\end{figure*}

\begin{figure*}[h!]
    \centering
    \includegraphics[width=.5\textwidth]{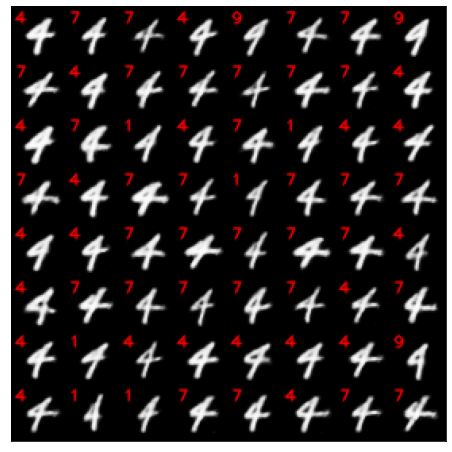}
    \caption{Annotator label $4$.}
\end{figure*}

% \begin{figure*}[h!]
%     \centering
%     \includegraphics[width=.5\textwidth]{images/mnistfailurescenarioexamples/28-2.png}
%     \caption{Annotator label $2$.}
% \end{figure*}

\begin{figure*}[h!]
    \centering
    \includegraphics[width=.5\textwidth]{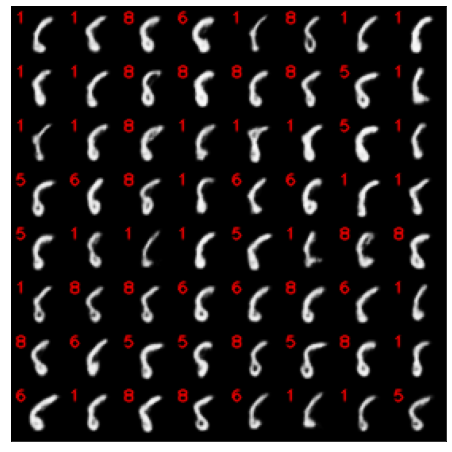}
    \caption{Annotator label $6$.}
\end{figure*}

\begin{figure*}[h!]
    \centering
    \includegraphics[width=.5\textwidth]{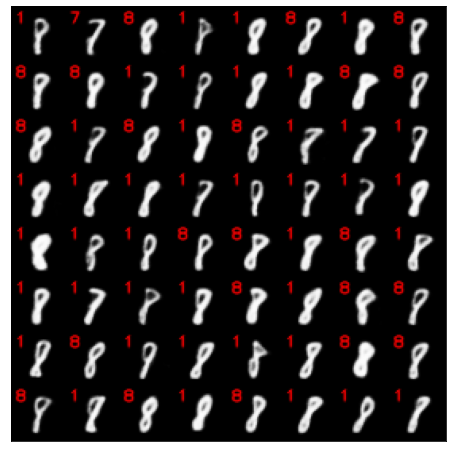}
    \caption{Annotator label $8$.}
\end{figure*}

\begin{figure*}[h!]
    \centering
    \includegraphics[width=.5\textwidth]{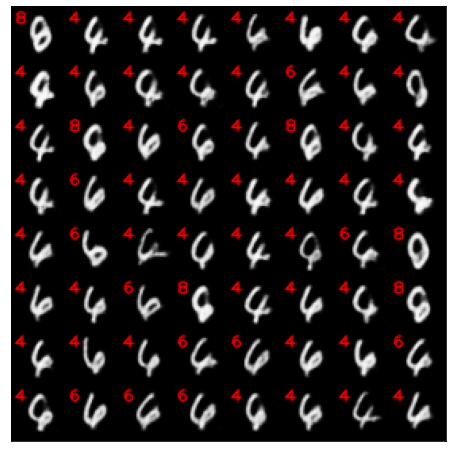}
    \caption{Annotator label $6$.}
\end{figure*}

\begin{figure*}[h!]
    \centering
    \includegraphics[width=.5\textwidth]{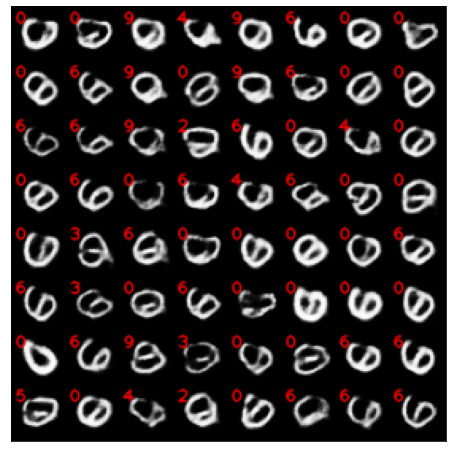}
    \caption{Annotator label $0$.}
\end{figure*}

\begin{figure*}[h!]
    \centering
    \includegraphics[width=.5\textwidth]{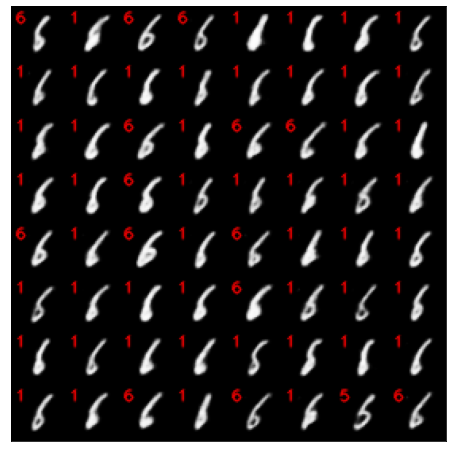}
    \caption{Annotator label $6$.}
\end{figure*}

\begin{figure*}[h!]
    \centering
    \includegraphics[width=.5\textwidth]{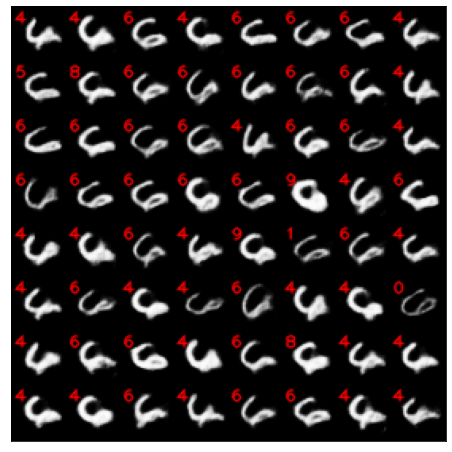}
    \caption{Annotator label $6$.}
\end{figure*}

\clearpage

\subsection{Additional samples from German signs misclassification regions}
\label{ap:additionalgermansigns}

We provide samples from all of the German signs misclassification regions. We provide the names of the class labels in figure \ref{fig:classlabelssigns}.  For each misclassification region, we indicate our assigned class label in the caption and the classifier predictions in the upper right hand corner of the image.

\begin{figure*}[h!]
\centering
\begin{tabular}{l|l}
ClassId & SignName \\
0 & Speed limit (20km/h) \\
1&Speed limit (30km/h) \\
2&Speed limit (50km/h) \\
3&Speed limit (60km/h) \\
4&Speed limit (70km/h) \\
5&Speed limit (80km/h) \\
6&End of speed limit (80km/h) \\
7&Speed limit (100km/h) \\
8&Speed limit (120km/h) \\
9&No passing \\
10&No passing for vehicles over 3.5 metric tons \\
11&Right-of-way at the next intersection \\
12&Priority road \\
13&Yield \\
14&Stop \\
15&No vehicles \\
16&Vehicles over 3.5 metric tons prohibited \\
17&No entry \\
18&General caution \\
19&Dangerous curve to the left \\
20&Dangerous curve to the right \\
21&Double curve \\
22&Bumpy road \\
23&Slippery road \\
24&Road narrows on the right \\
25&Road work \\
26&Traffic signals \\
27&Pedestrians \\
28&Children crossing \\
29&Bicycles crossing \\
30&Beware of ice/snow \\
31&Wild animals crossing \\
32&End of all speed and passing limits \\
33&Turn right ahead \\
34&Turn left ahead \\
35&Ahead only \\
36&Go straight or right \\
37&Go straight or left \\
38&Keep right \\
39&Keep left \\
40&Roundabout mandatory \\
41&End of no passing \\
42&End of no passing by vehicles over 3.5 metric tons \\
\end{tabular}
\caption{German signs class labels.}
\label{fig:classlabelssigns}
\end{figure*}

\begin{figure*}[h!]
    \centering
    \includegraphics[width=.5\textwidth]{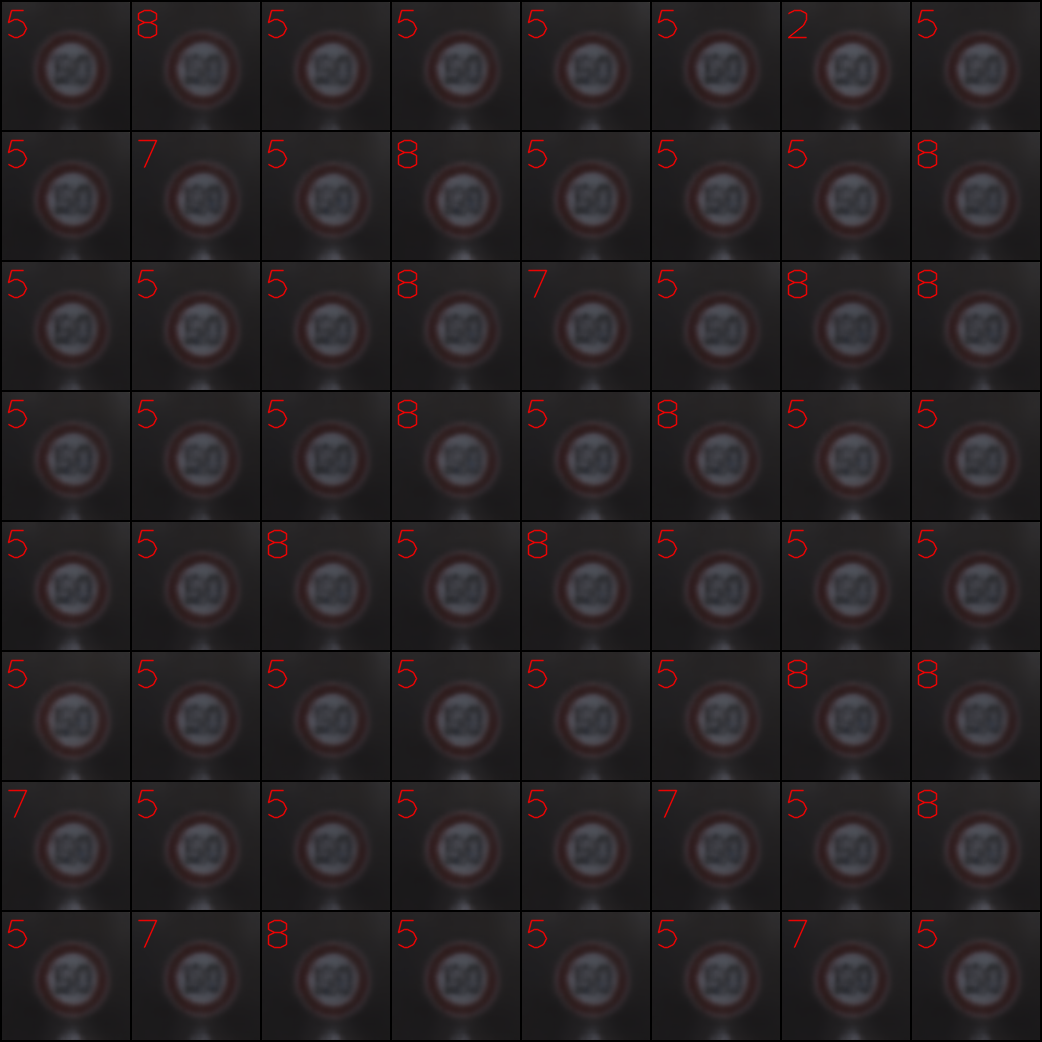}
    \caption{Annotator label $7$.}
\end{figure*}

\begin{figure*}[h!]
    \centering
    \includegraphics[width=.5\textwidth]{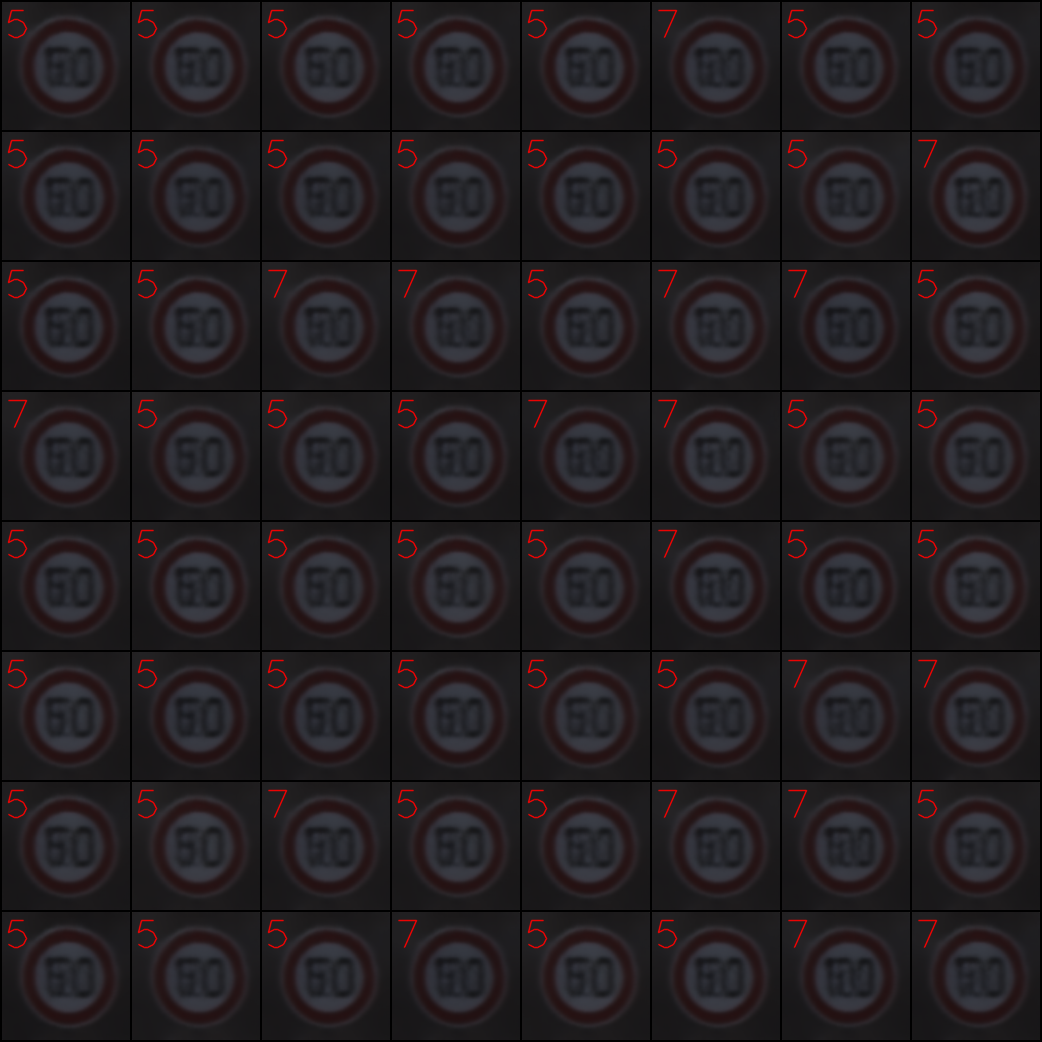}
    \caption{Annotator label $2$.}
\end{figure*}

\begin{figure*}[h!]
    \centering
    \includegraphics[width=.5\textwidth]{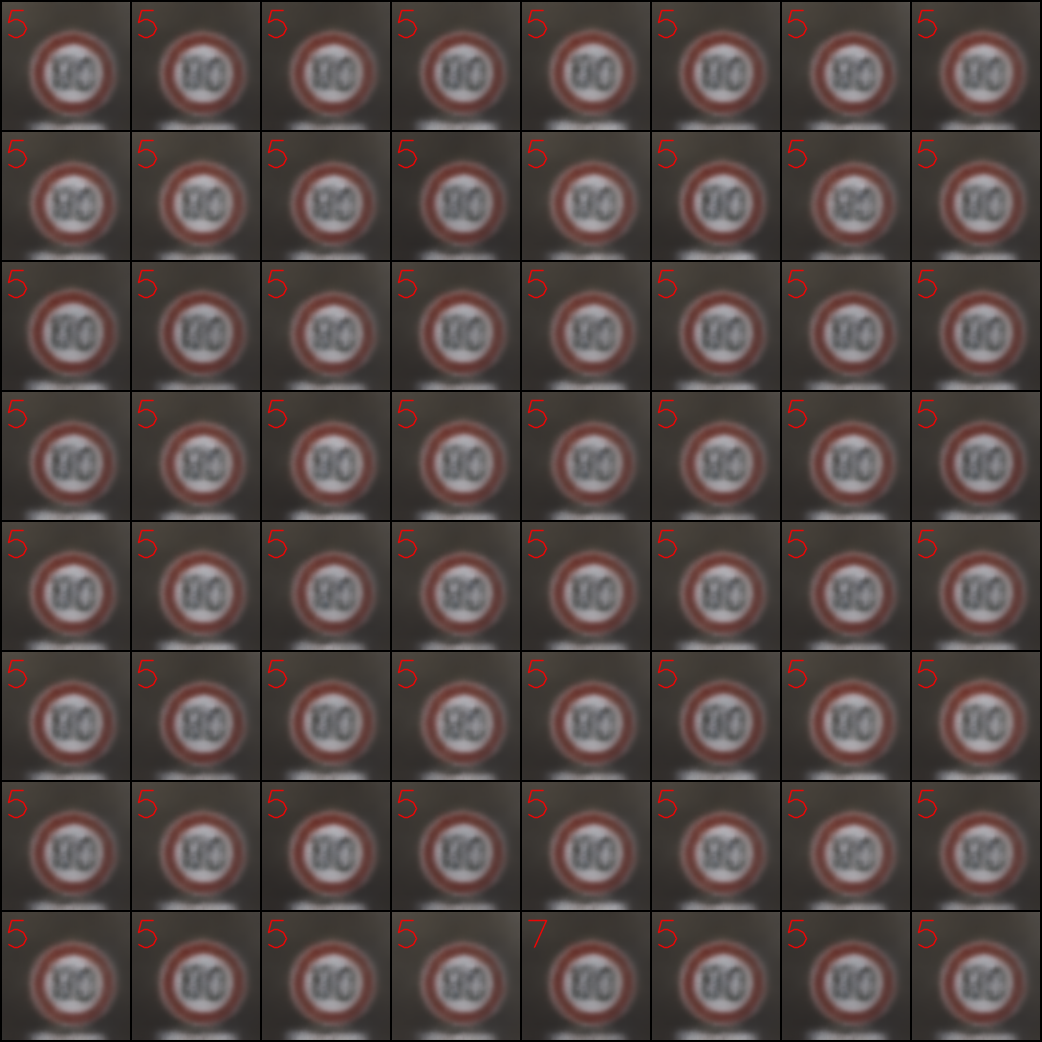}
    \caption{Annotator label $7$.}
\end{figure*}

\begin{figure*}[h!]
    \centering
    \includegraphics[width=.5\textwidth]{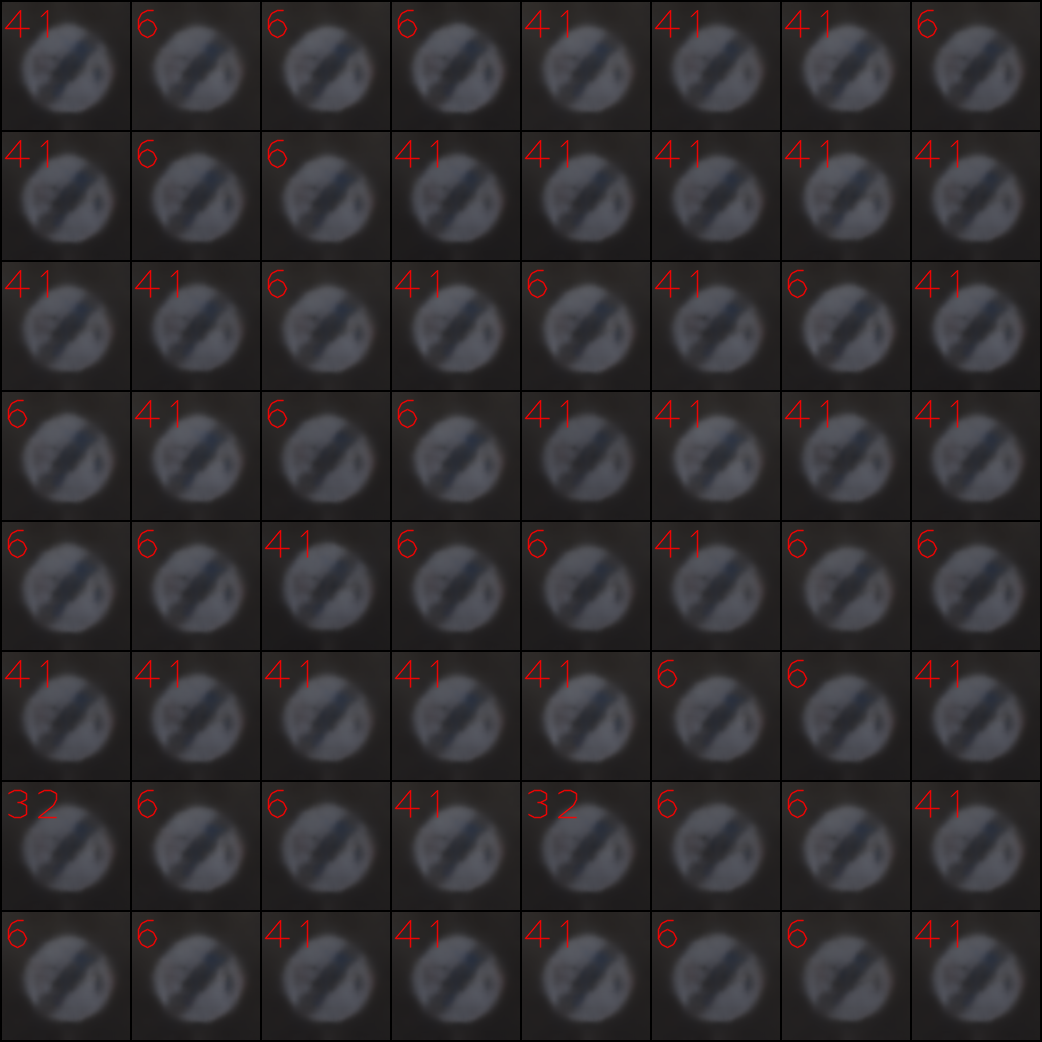}
    \caption{Annotator label $41$.}
\end{figure*}

\begin{figure*}[h!]
    \centering
    \includegraphics[width=.5\textwidth]{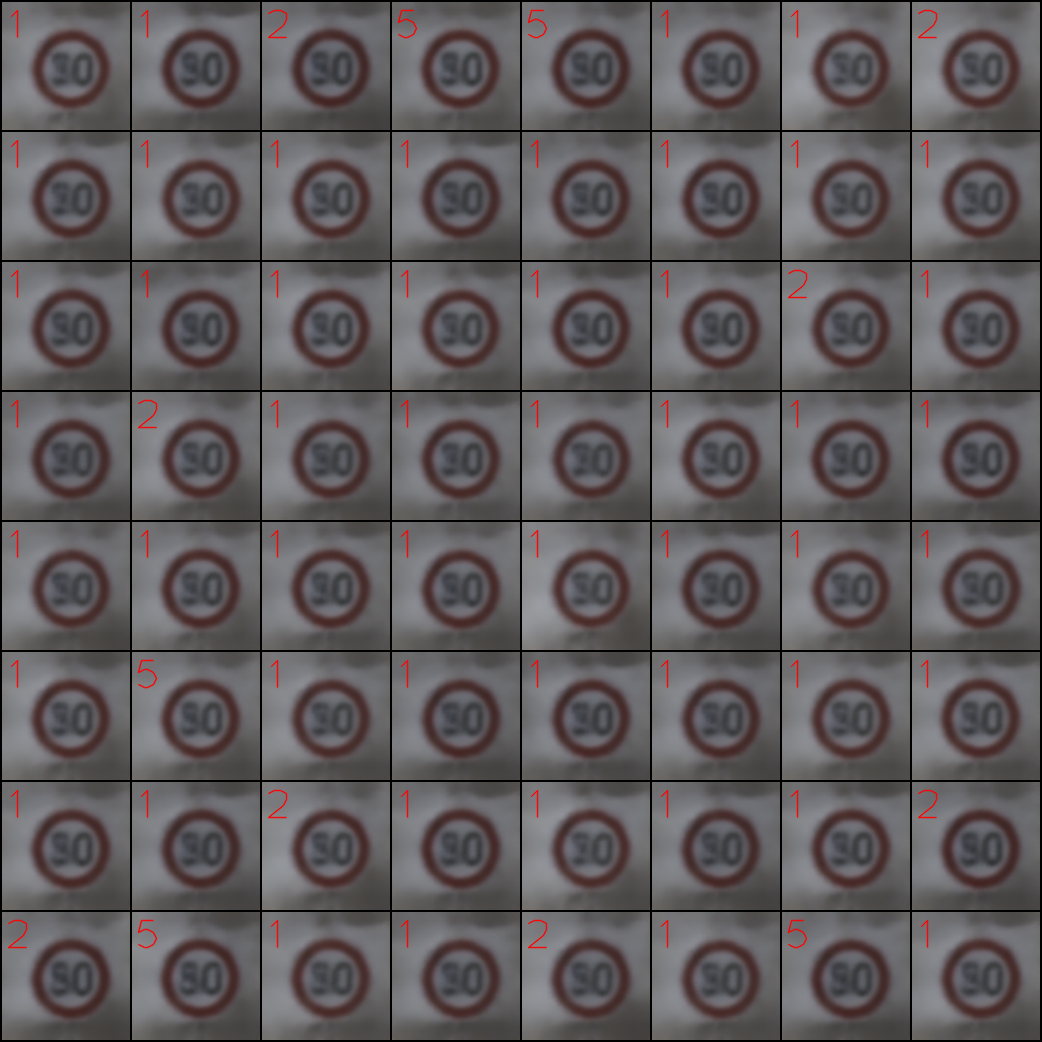}
    \caption{Annotator label $1$.}
\end{figure*}

\begin{figure*}[h!]
    \centering
    \includegraphics[width=.5\textwidth]{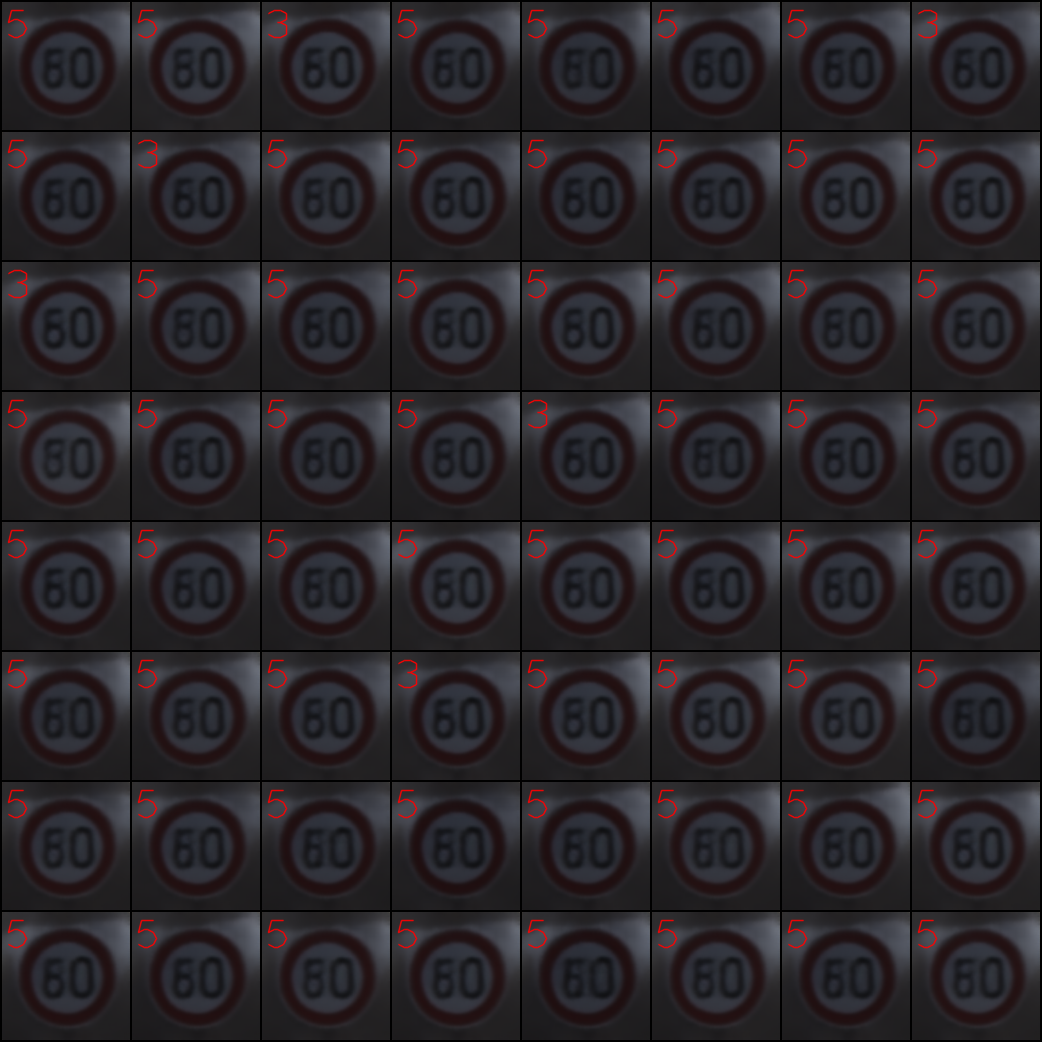}
    \caption{Annotator label $2$.}
\end{figure*}

\begin{figure*}[h!]
    \centering
    \includegraphics[width=.5\textwidth]{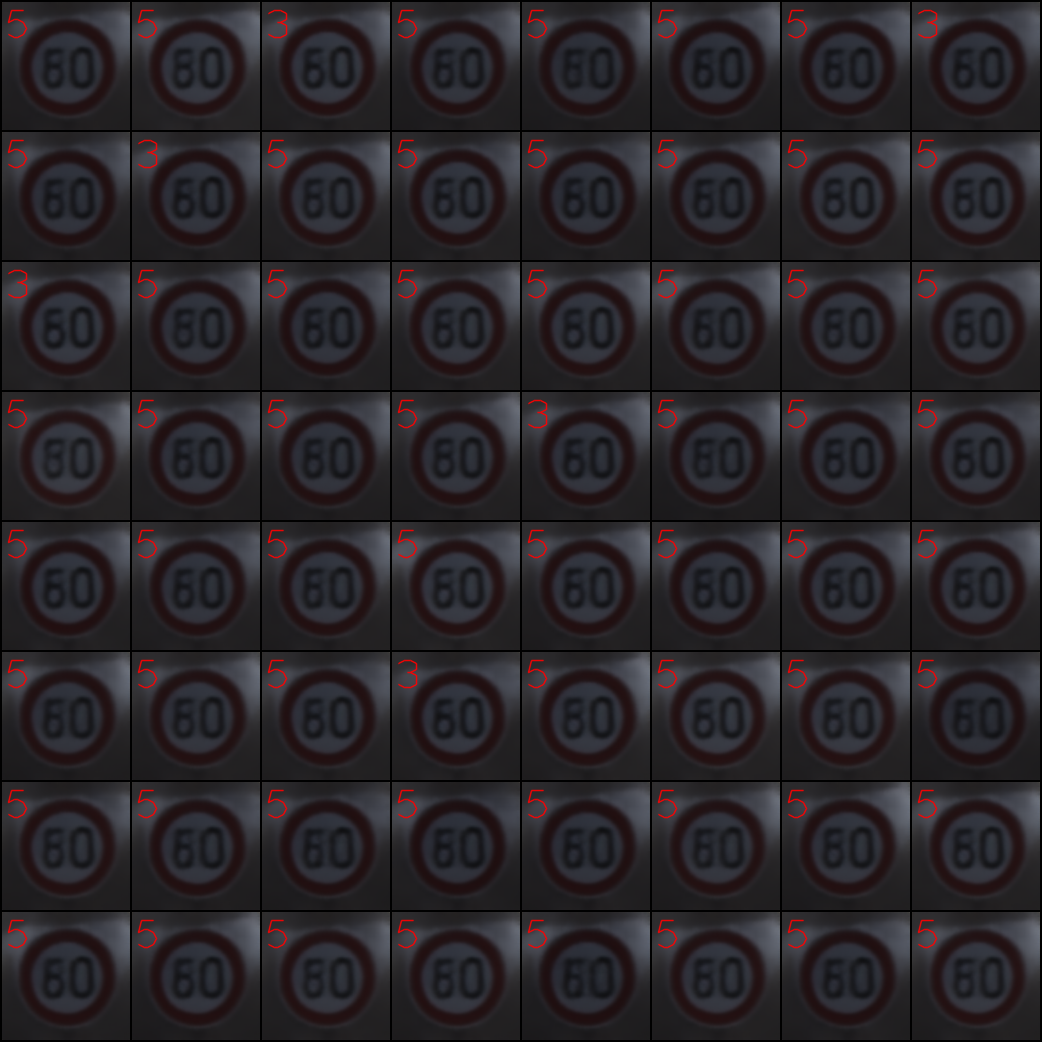}
    \caption{Annotator label $2$.}
\end{figure*}

\begin{figure*}[h!]
    \centering
    \includegraphics[width=.5\textwidth]{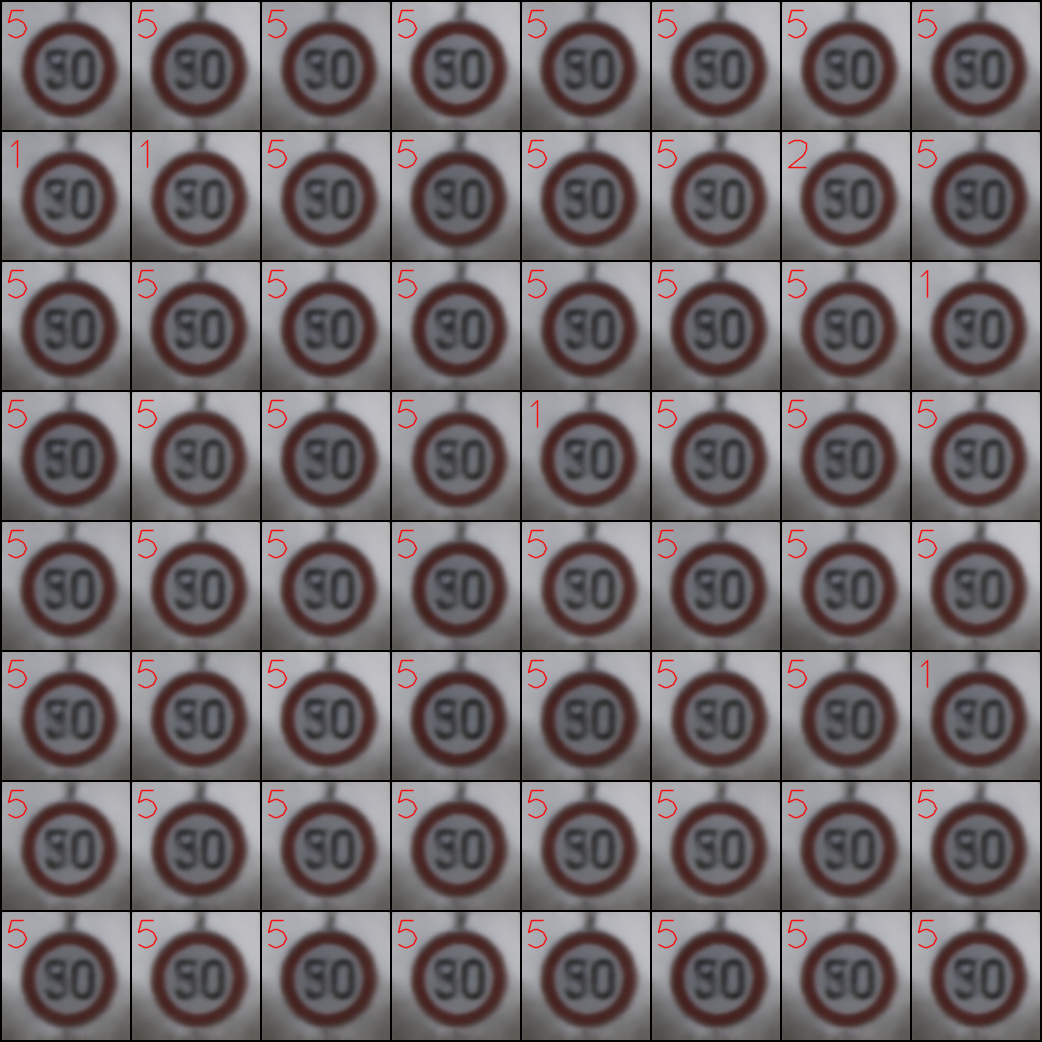}
    \caption{Annotator label $1$.}
\end{figure*}

\begin{figure*}[h!]
    \centering
    \includegraphics[width=.5\textwidth]{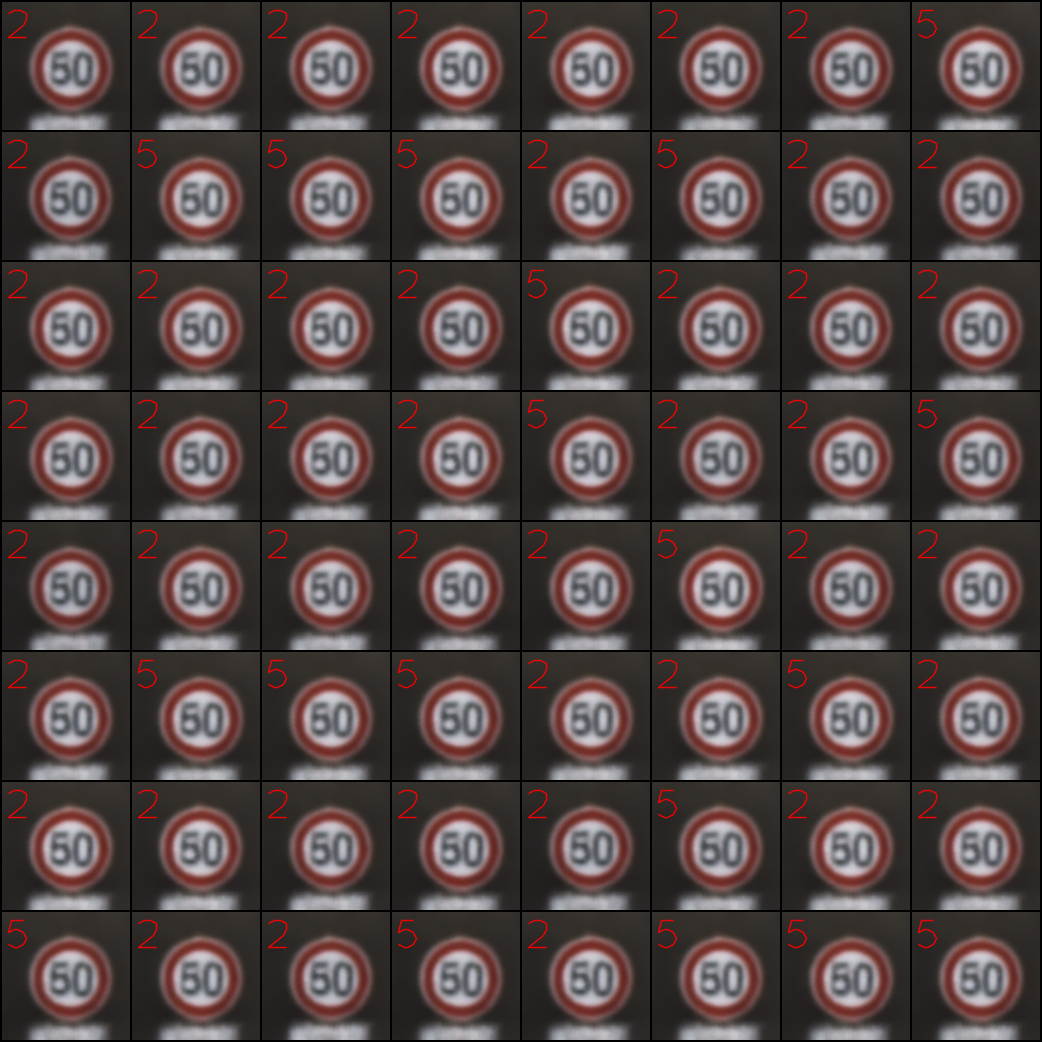}
    \caption{Annotator label $2$.}
\end{figure*}

\clearpage

\subsection{Additional samples from SVHN misclassification regions}
\label{sec:svhnfailurescenarios}

We provide additional samples from each of the SVHN misclassification regions.  The digit in the upper left hand corner is the classifier predicted label.  The caption includes the Ground Truth worker labels.

\begin{figure*}[h!]
    \centering
    \includegraphics[width=.5\textwidth]{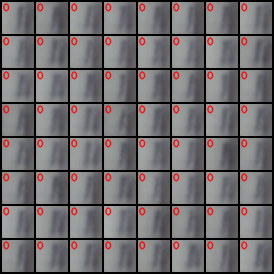}
    \caption{Annotator label $1$.}
\end{figure*}

\begin{figure*}[h!]
    \centering
    \includegraphics[width=.5\textwidth]{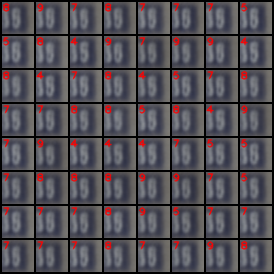}
    \caption{Annotator label $5$.}
\end{figure*}

\begin{figure*}[h!]
    \centering
    \includegraphics[width=.5\textwidth]{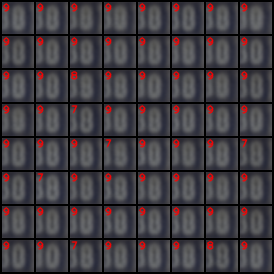}
    \caption{Annotator label $8$.}
\end{figure*}

\begin{figure*}[h!]
    \centering
    \includegraphics[width=.5\textwidth]{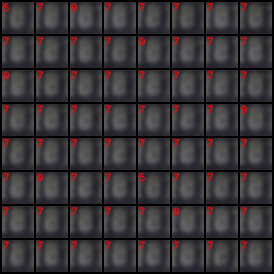}
    \caption{Annotator label $0$.}
\end{figure*}

\begin{figure*}[h!]
    \centering
    \includegraphics[width=.5\textwidth]{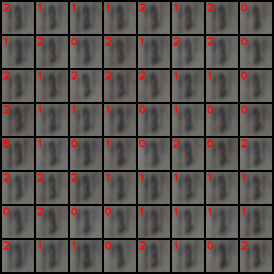}
    \caption{Annotator label $3$.}
\end{figure*}

\begin{figure*}[h!]
    \centering
    \includegraphics[width=.5\textwidth]{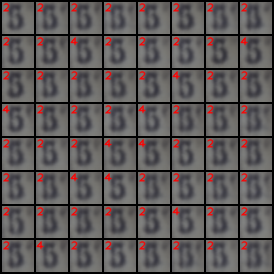}
    \caption{Annotator label $5$.}
\end{figure*}

\end{document}